\documentclass{article}

\usepackage{amsmath}
\usepackage{amssymb}
\usepackage{graphicx}
\usepackage{subcaption}
\usepackage{multirow}
\usepackage{xcolor} 

    \usepackage[preprint,nonatbib]{neurips_2025}

\usepackage[utf8]{inputenc} 
\usepackage[T1]{fontenc}    
\usepackage{hyperref}       
\usepackage{url}            
\usepackage{booktabs}       
\usepackage{amsfonts}       
\usepackage{nicefrac}       
\usepackage{microtype}      
\usepackage{xcolor}         

\title{Qwen Look Again: Guiding Vision-Language Reasoning Models to Re-attention Visual Information}

\author{
  \textbf{Xu Chu\textsuperscript{1,†}},
  \textbf{Xinrong Chen\textsuperscript{1,†}},
  \textbf{Guanyu Wang\textsuperscript{1,†}},
  \textbf{Zhijie Tan\textsuperscript{1}}
\\
  \textbf{Kui Huang\textsuperscript{2}},
  \textbf{Wenyu Lv\textsuperscript{2}},
  \textbf{Tong Mo\textsuperscript{1}},
  \textbf{Weiping Li\textsuperscript{1,*}}
\\
\\
  \textsuperscript{1}Peking University, Beijing, China, \textsuperscript{2}Baidu Inc., Beijing, China
\\
  \href{mailto:chuxu@stu.pku.edu.cn}{chuxu@stu.pku.edu.cn},
  \href{mailto:chenxinrong23@stu.pku.edu.cn}{chenxinrong23@stu.pku.edu.cn},
  \href{mailto:wangguanyu@stu.pku.edu.cn}{wgy2023@stu.pku.edu.cn}
\\
  \href{mailto:besttangent@stu.pku.edu.cn}{besttangent@stu.pku.edu.cn},
  \href{mailto:huangkui01@baidu.com}{huangkui01@baidu.com},
  \href{mailto:lvwenyu01@baidu.com}{lvwenyu01@baidu.com}
\\
  \href{mailto:motong@ss.pku.edu.cn}{motong@ss.pku.edu.cn},
  \href{mailto:wpli@ss.pku.edu.cn}{wpli@ss.pku.edu.cn}
\\
  \small{
    \textsuperscript{†}Equal contribution
    \hspace{2em}
    \textsuperscript{*}Corresponding author
  }
}

\begin{document}

\maketitle

\begin{abstract}
Inference time scaling drives extended reasoning to enhance the performance of Vision-Language Models (VLMs), thus forming powerful Vision-Language Reasoning Models (VLRMs). However, long reasoning dilutes visual tokens, causing visual information to receive less attention and may trigger hallucinations. Although introducing text-only reflection processes shows promise in language models, we demonstrate that it is insufficient to suppress hallucinations in VLMs. To address this issue, we introduce \textbf{Qwen-\underline{L}ook\underline{A}gain (Qwen-LA)}, a novel VLRM designed to mitigate hallucinations by incorporating a vision-text reflection process that guides the model to re-attention visual information during reasoning. We first propose a reinforcement learning method \textbf{\underline{B}alanced \underline{R}eflective \underline{P}olicy \underline{O}ptimization (BRPO)}, which guides the model to decide when to generate vision-text reflection on its own and balance the number and length of reflections. Then, we formally prove that VLRMs lose attention to visual tokens as reasoning progresses, and demonstrate that supplementing visual information during reflection enhances visual attention. Therefore, during training and inference, \textbf{Visual Token COPY} and \textbf{Visual Token ROUTE} are introduced to force the model to re-attention visual information at the visual level, addressing the limitations of text-only reflection. Experiments on multiple visual QA datasets and hallucination metrics indicate that Qwen-LA achieves leading accuracy performance while reducing hallucinations. Our code is available at: \url{https://github.com/Liar406/Look_Again}.
\end{abstract}

\section{Introduction}
Large Reasoning Language Models (LRLM) are driven by inference time scaling to conduct extended reasoning. These models achieve significant performance improvements across multiple tasks, including mathematics, coding, and creative writing, by generating longer decoded text~\cite{chen2025towards,jaech2024openai, qwq32b,guo2025deepseek,aggarwal2025l1,wu2025effectively}. In the field of Vision-Language Models (VLMs), recent research explores the integration of extended reasoning into VLMs to enhance their performance, introducing Vision-Language Reasoning Models (VLRMs) with promising results~\cite{liu2025visual,huang2025vision,team2025kimi}. However, additional reasoning may introduce extra hallucinations, leading to unexpected errors.
\begin{figure*}[t]
    \centering
    \begin{minipage}[t]{0.48\textwidth}
        \begin{subfigure}[b]{0.48\textwidth}
            \includegraphics[width=\textwidth]{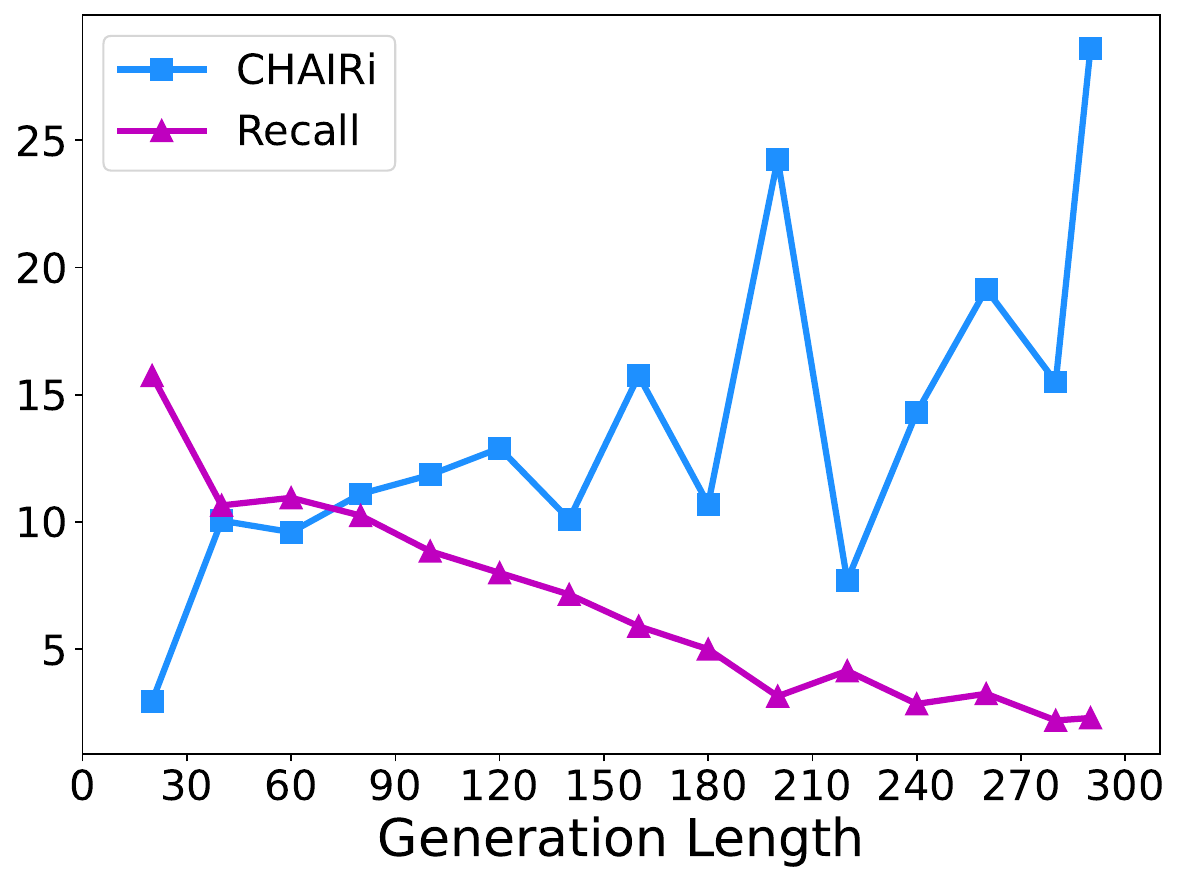}

            \caption{$\text{CHAIR}_\text{i}$ and Recall during generation.}
            \label{fig:hal_org}
        \end{subfigure}
        \hfill
        \begin{subfigure}[b]{0.48\textwidth}
            \includegraphics[width=\textwidth]{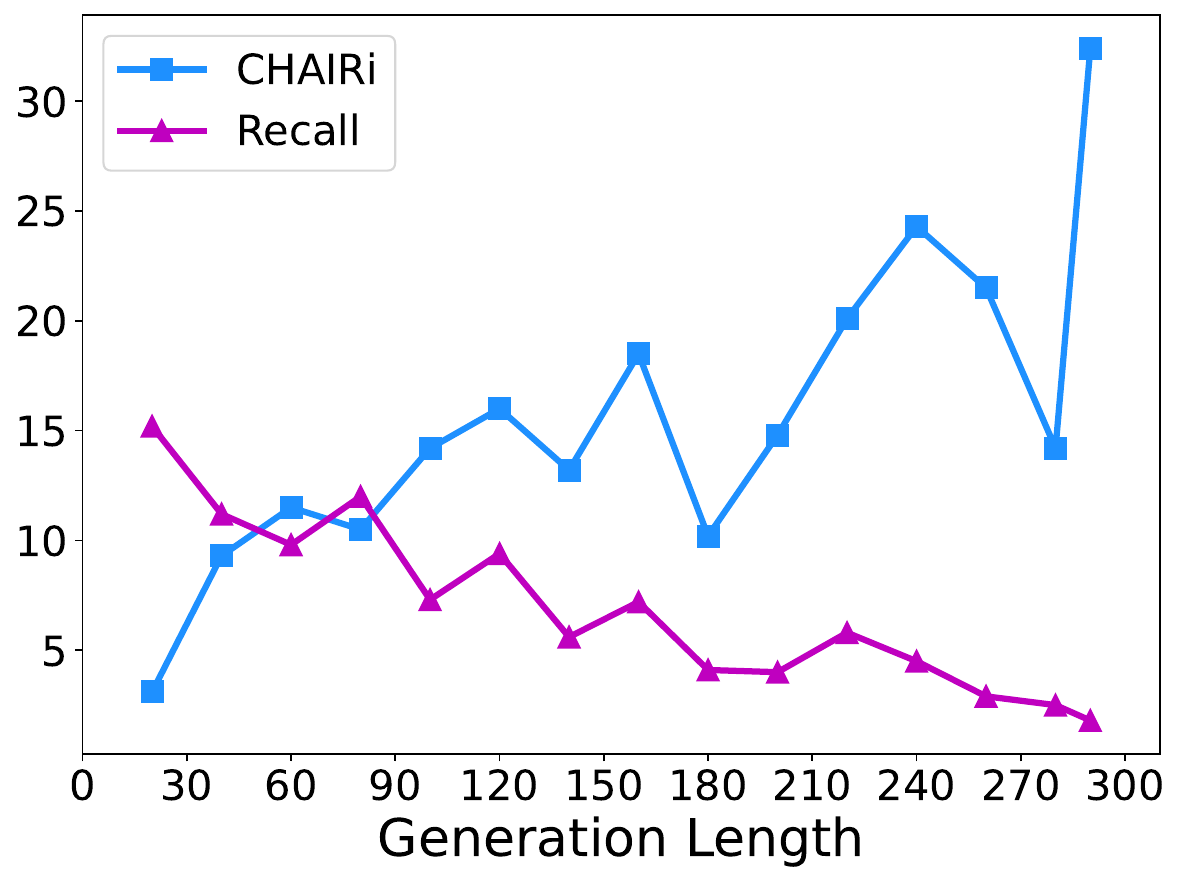}

            \caption{$\text{CHAIR}_\text{i}$ and Recall with text-only reflection.}
            \label{fig:hal_text}
        \end{subfigure}
        \vspace{0.1em}
        \caption{Analysis of hallucination metrics on 500 randomly selected examples from MSCOCO~\cite{lin2014microsoft} dataset. As the generation length increases, both standard generation and text-only reflection show increased $\text{CHAIR}_\text{i}$ and decreased Recall.}
        \label{fig:hal}
    \end{minipage}
    \hfill
    \begin{minipage}[t]{0.48\textwidth}
        \begin{subfigure}[b]{0.48\textwidth}
            \includegraphics[width=\textwidth]{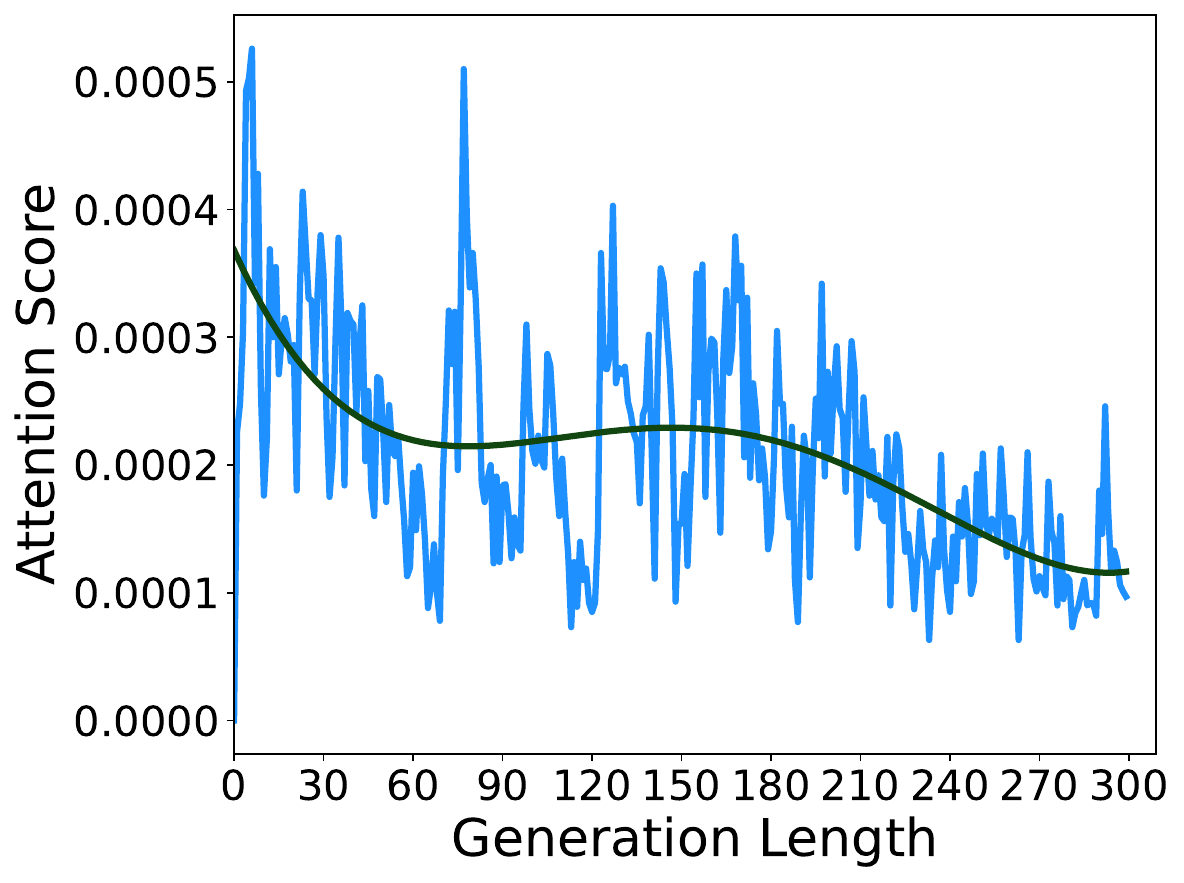}

            \caption{Attention weights during generation.}
            \label{fig:org_attn}
        \end{subfigure}
        \hfill
        \begin{subfigure}[b]{0.48\textwidth}
            \includegraphics[width=\textwidth]{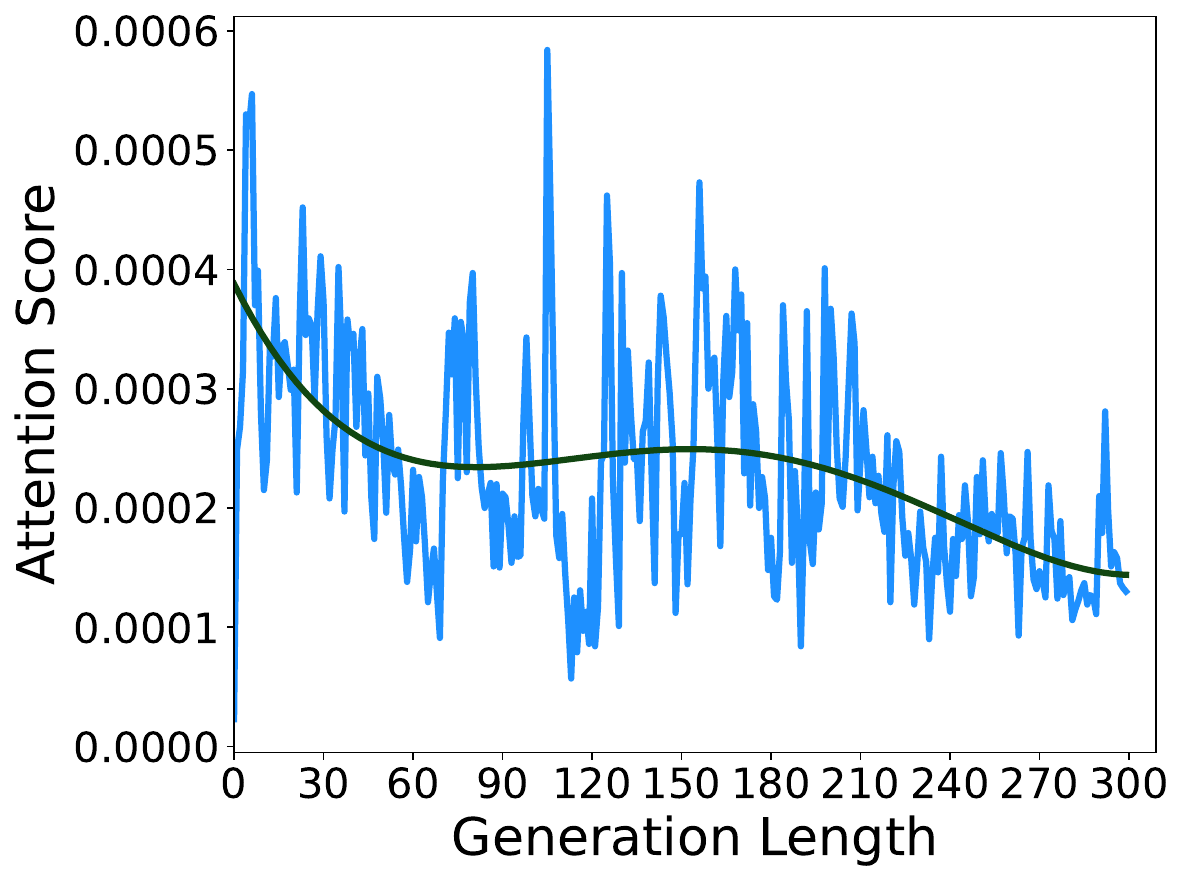}
 
            \caption{Attention weights with text-only reflection.}
            \label{fig:text_attn}
        \end{subfigure}
        \vspace{0.1em}
        \caption{Analysis of visual attention patterns on 500 randomly selected examples from MSCOCO~\cite{lin2014microsoft} dataset. As generation progresses, attention weights of visual tokens decrease. This indicates that as tokens are generated, the focus on visual information diminishes.}
        \label{fig:attention}
    \end{minipage}
    \vspace{0.3em}  
\end{figure*}

Hallucinations in VLMs can be defined as generating content that is irrelevant or contradictory to the facts in the image~\cite{favero2024multi}. Multiple studies indicate that hallucinations in VLMs are related to the language priors accumulated during the VLM's generation process~\cite{favero2024multi, zhou2023analyzing, li2025hidden}. As the model generates responses, the textual content gradually dilutes the visual context, leading to grammatically coherent but visually ungrounded content. This becomes more prominent in VLRMs with longer generation. As shown in Figure~\ref{fig:hal_org}, during the generation process, LLaVA-CoT~\cite{xu2411llava}, as a VLRM, exhibits an intensified hallucination (hallucination metric $\text{CHAIR}_\text{i}$) and decreased Recall. An intuitive idea is to use text prompts such as "Look at the image again" to drive the model to reflect~\cite{madaan2023self,zhang-etal-2024-self-contrast} and suppress hallucinations. However, we find that despite the model reflecting through text-only prompts, as shown in Figure~\ref{fig:hal_text}, hallucinations do not decrease. Figure~\ref{fig:org_attn} further analyzes the relationship between generation length and visual attention from an attention perspective, showing that the average attention weights of visual tokens decrease as generation progresses. Figure~\ref{fig:text_attn} indicates that guiding the model to explicitly output a reflection process through text-only prompts does not increase the attention weights of visual tokens. Our analysis shows that \textbf{(1)} the long reasoning generation of VLRMs reduce visual attention, and \textbf{(2)} even when the model is explicitly prompted to review the input image and reflect on earlier reasoning processes, it may not actually re-attend to visual tokens.

In this paper, we present \textbf{Qwen-\underline{L}ook\underline{A}gain (Qwen-LA)}, a novel VLRM for hallucination suppression and accuracy improvement. Qwen-LA uses reinforcement learning to guide the model to spontaneously introduce vision-text reflection processes and force it to re-attention visual information to correct errors in reasoning. Specifically, we first propose \textbf{\underline{B}alanced \underline{R}eflective \underline{P}olicy \underline{O}ptimization (BRPO)}, a rule-based reinforcement learning method that trains Qwen2.5-VL-7B-Instruct~\cite{bai2025qwen2} to generate reasoning processes with n (n$\geq$1) reflections on its own, resulting in Qwen-Zero. Notably, we observe that the model spontaneously generates multiple reflections at different positions in the response through reinforcement learning, and these reflections become more concise as training iterations progress. Qwen-Zero is used to construct Qwen-Zero-40k, a reasoning dataset with reflections, which undergoes human verification and correction. Then, we formally prove that as tokens are generated during the generation process, the VLM's attention to visual information decreases. We further demonstrate that enhancing visual information during model generation can increase visual attention. Therefore, we improve and train Qwen2.5-VL-7B-Instruct by parallelly introducing two methods, \textbf{\underline{V}isual \underline{T}oken \underline{C}OPY (VTC)} and \textbf{\underline{V}isual \underline{T}oken \underline{R}OUTE (VTR)}, to force the model to re-attention visual information. When the model generates a reflection process (start with <REFLECTION> token), VTC simply copies the complete visual tokens of the input image to the beginning of the reflection process. VTR, on the other hand, routes visual tokens with higher attention weights to the beginning of the reflection process based on contextual attention distribution. Both methods use the Qwen-Zero-40k dataset and undergo full-parameter supervised fine-tuning, resulting in Qwen-LA-COPY and Qwen-LA-ROUTE respectively. We evaluate Qwen-LA-COPY and Qwen-LA-ROUTE's accuracy metrics on multiple visual QA datasets and assess various hallucination metrics on multiple hallucination datasets. Experimental results demonstrate that our models achieve leading performance in both accuracy and hallucination metrics. Our contributions are as follows:
\begin{itemize}
    \item We discover that as VLRM generates lengthy reasoning, hallucinations intensify with increasing generation length. Text-only reflection do not mitigate this issue. Furthermore, we formally prove that the occurrence of hallucinations is related to reasoning length.
    \item We propose Qwen-LA, a novel VLRM that suppresses hallucinations and improves accuracy. The proposed reinforcement learning method BRPO, guides the model to spontaneously perform reasoning with vision-text reflection. Visual Token COPY and Visual Token ROUTE are introduced to force the model to re-attention visual information.
    \item Our experiments on multiple visual QA datasets and hallucination metrics demonstrate that Qwen-LA effectively suppresses hallucinations and achieves leading performance.
\end{itemize}

\section{Related Works}
\subsection{Hallucination in VLMs}
Although VLM hallucination is considered multifaceted~\cite{liu2024survey}, a key cause stems from language priors overwhelming visual context, which has been studied from the perspective of attention patterns~\cite{huang2024opera, LiuZC24}. Wang et al.~\cite{wang2024picture} observe that even with access to image data, VLMs can sometimes respond based on textual information and hallucination rather than directly utilizing visual content. Huang et al.~\cite{huang2024opera} observe that hallucination may arise when VLMs over-trust summary tokens in context while neglecting image tokens. Favero et al.~\cite{favero2024multi} and Li et al~\cite{li2025hidden} further discover that VLMs' reliance on vision decreases as more tokens are generated, indicating that visual information becomes diluted and ignored during the model's autoregressive generation process. Although recent studies mitigate hallucination through visual prompting or adjusting visual token weights~\cite{LiuZC24, yu2025introducing, interleaved2025}, most of them artificially decide when to introduce these methods, such as at the beginning of all paragraphs. In our research, we find that VLMs can be guided through reinforcement learning to spontaneously decide when to re-attend to visual information, without relying on human priors.

\subsection{Inference Time Scaling}
Extensive research indicates that inference time scaling can improve LLMs' reasoning performance~\cite{xie2024monte, chu2025domaino1s, guo2025deepseek, team2025kimi}. To optimize reasoning processes represented by Chain-of-Thought (CoT), one category of methods obtains optimal reasoning paths from the solution space through tree search, either by fine-tuning models~\cite{xie2024monte, chen2024step, yuan2024advancing} or directly exploring optimal reasoning paths during inference~\cite {Labrak2024BioMistralAC,xu2411llava,chu2025domaino1s}. Another category of methods, like DeepSeek-R1~\cite{guo2025deepseek}, stimulates models' inherent reasoning capabilities through reinforcement learning and observes the emergence of reflection processes. Inspired by this, many researchers introduce reinforcement learning to train VLRMs~\cite{liu2025visual,huang2025vision,team2025kimi}. However, both LRLMs and VLRMs face excessive overhead and hallucinations caused by lengthy reasoning processes~\cite{chen2025towards, liao2025longperceptualthoughts}.

\begin{figure*}[t]
    \centering
    \includegraphics[width=\textwidth]{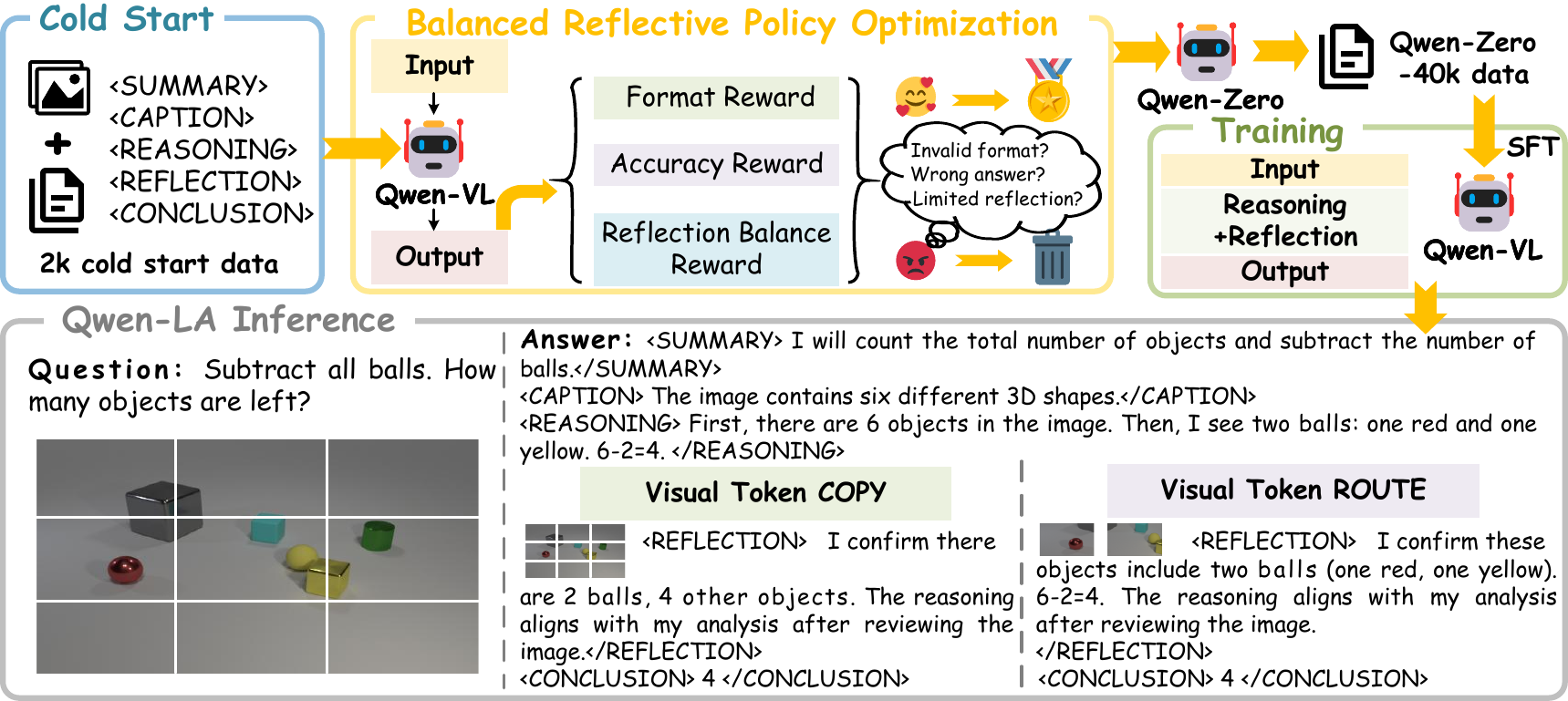}
    \caption{The framework of Qwen-LookAgain. The collected cold-start data is used to fine-tune Qwen2.5-VL-Instruct~\cite{bai2025qwen2}. Then, Balanced Reflective Policy Optimization (BRPO) reward Qwen2.5-VL-Instruct spontaneously generates more accurate, standard answers and more powerful reflection processes, resulting in Qwen-Zero. Next, Qwen-Zero's distilled data is used to SFT Qwen2.5-VL-Instruct, producing Qwen-LA. During SFT and inference stages, Visual Token COPY or Visual Token ROUTE guides the model to re-attention the visual information during the reflection process.}
    \label{fig:qwen_la}
\end{figure*}
\section{Methodology}
In this section, we propose Qwen-LookAgain (Qwen-LA), a novel VLRM based on reinforcement learning and visual re-attention, with its framework shown in Figure~\ref{fig:qwen_la}. Qwen-LA spontaneously generates multiple reflections through the proposed BRPO (Section~\ref{section:BRPO}) and guides the model to re-attention visual information (Section~\ref{Visual Re-attention}) through Visual Token COPY or Visual Token ROUTE, thereby suppressing hallucinations.
\subsection{Preliminaries}
\textbf{Rule-based Reinforcement Learning}. Rule-based RL aims to improve language models' performance in tasks with clear correct answers (such as mathematics and programming). Unlike traditional Reinforcement Learning from Human Feedback (RLHF)~\cite{ouyang2022training}, Rule-based RL does not rely on training complex reward models, but utilizes direct, rule-based validation functions to evaluate output correctness. This approach simplifies the reward mechanism through explicit rule sets while ensuring model outputs highly align with the inherent correctness criteria of tasks. Rule-based RL is particularly effective in scenarios lacking annotated data. For example, the Group Relative Policy Optimization (GRPO)~\cite{shao2024deepseekmath} framework used to train DeepSeek-R1-Zero~\cite{guo2025deepseek} eliminates dependence on supervised data and human preference data, guiding models to spontaneously explore reasoning paths from the solution space.

GRPO is based on a simple yet effective rule: directly comparing the relative merits of multiple candidate answers for the same problem. Specifically, for a given question $q$, GRPO first uses the current policy $\pi_{\text{old}}$ to generate $G$ different responses $\{o_1, o_2, \dots, o_G\}$ and obtains corresponding rewards $\{r_1, r_2, \dots, r_G\}$. GRPO calculates the relative advantage of each answer through intra-group normalization:

\begin{equation}
A_i = \frac{r_i - \text{mean}(\{r_1, \dots, r_G\})}{\text{std}(\{r_1, \dots, r_G\})},
\end{equation}

where $A_i$ represents the degree of advantage of the $i$-th answer relative to other answers within the group. Based on these relative advantage values, the GRPO objective function can be expressed as:

{\small
\begin{equation}
\label{eq:grpo}
\mathcal{L}_{GRPO}(\theta) = \frac{1}{G} \sum_{i=1}^{G} \left(\min \left( \frac{\pi_{\theta}(o_i|q)}{\pi_{\text{old}}(o_i|q)} A_i, \text{clip}\left(\frac{\pi_{\theta}(o_i|q)}{\pi_{\text{old}}(o_i|q)}, 1-\epsilon, 1+\epsilon\right) A_i \right)- \beta\mathbb{D}_{KL} (\pi_{\theta}||\pi_{ref})\right),
\end{equation}
}

\begin{equation}
\mathbb{D}_{KL} (\pi_{\theta}||\pi_{ref}) = \frac{\pi_{ref}(o_i|q)}{\pi_{\theta}(o_i|q)} - \log \frac{\pi_{ref}(o_i|q)}{\pi_{\theta}(o_i|q)} - 1,
\end{equation}

where $\epsilon$ and $\beta$ control the clipping range and the intensity of KL divergence penalty respectively. $\pi_{ref}$ is the reference policy, used to prevent the optimized policy $\pi_{\theta}$ from deviating too far and causing catastrophic forgetting.

\textbf{Vision-Language Models}. The input of Vision-Language Models (VLMs) includes text prompts $x$ and image prompts $c$. The text prompts contain $L_x$ tokens, and the image prompts are processed by visual encoders into $L_c$ visual tokens. VLMs generate output sequences $y$ of length $L_y$ in an autoregressive manner, with the conditional probability of the entire process expressed as $p(y|x,c)=\prod_{t=1}^{L_y} p(y_t \mid y_{<t}, x, c)$, where $y_{<t} = [y_1,\dots,y_{t-1}]$ represents the generated tokens.

\subsection{Balanced Reflective Policy Optimization}
\label{section:BRPO}
Inspired by GRPO, we propose Balanced Reflective Policy Optimization (BRPO). Specifically, BRPO employs a rule-based reward system consisting of three types of rewards:

    $\bullet$ Format reward: We enforce the model to respond in a specified format, including a summary process (<SUMMARY></SUMMARY>), an image caption process (<CAPTION></CAPTION>), step-by-step reasoning and thinking process (<REASONING></REASONING>), reflection process (<REFLECTION></REFLECTION>), and answer summarization process (<CONCLUSION></CONCLUSION>). The relative order of all processes except reflection is restricted, and no format overlapping or nesting relationships are allowed. The number and relative positions of reflection processes are unrestricted but similarly cannot overlap or nest.
    
    $\bullet$ Accuracy reward: Evaluates whether the final answer in the response is correct. Uses regular expressions to check if the content within <CONCLUSION></CONCLUSION> matches the correct answer, rewarding if consistent.
    
    $\bullet$ Reflection balance reward: Encourages models to maintain a balance between reflection count and total length when generating answers. When reflection count is low, longer reflections are allowed; when reflection count increases, reflection length is limited. The reward is formalized as $1 - \frac{|\frac{L_{r_{total}}} {N_r} - \lambda|}{\lambda}$, where $N_{r}$ is the number of reflection occurrences (i.e., occurrences of <REFLECTION></REFLECTION>), $L_{r_{total}}$ is the total number of tokens in all reflection processes, and $\lambda$ is the ideal average length for a single reflection.


With the reward mechanism determined, BRPO's objective function aligns with~(\ref{eq:grpo}). We apply the BRPO framework to train Qwen2.5-VL-7B-Instruct~\cite{bai2025qwen2}, using 10k data collected and mixed from Math~\cite{qiao2024we}, MathVision~\cite{wang2024measuring}, Polymath~\cite{gupta2024polymath}, SceMQA~\cite{liang2024scemqa}, and Geometry3K~\cite{lu2021inter}. To accelerate model convergence, we construct 2k cold-start data from LLaVA-CoT-100k~\cite{xu2411llava} and initially cold-start Qwen2.5-VL-7B-Instruct. The cold-start data inserts a reflection process between reasoning and conclusion processes (using GPT-4o~\cite{jaech2024openai} and manual verification to ensure process correctness), containing only \textbf{one} reflection in the cold-start data. The model trained through BRPO is called Qwen-Zero, with experimental setup details and results shown in Appendix~\ref{appendix:BRPO Experimental Details}. Notably, we observe two novel and interesting phenomena during training:

    $\bullet$ Even though the cold-start data contains only \textbf{one} reflection process, the model gradually produces \textbf{multiple} reflections under reward, and the total reflection length $L_{r_{total}}$ decreases as training epochs increase. This indicates that reflections become more numerous but concise, with the model automatically avoiding overthinking. As shown in Figure~\ref{fig:length_with_times}.
    
\begin{figure*}[t]
    \centering
    \includegraphics[width=\textwidth]{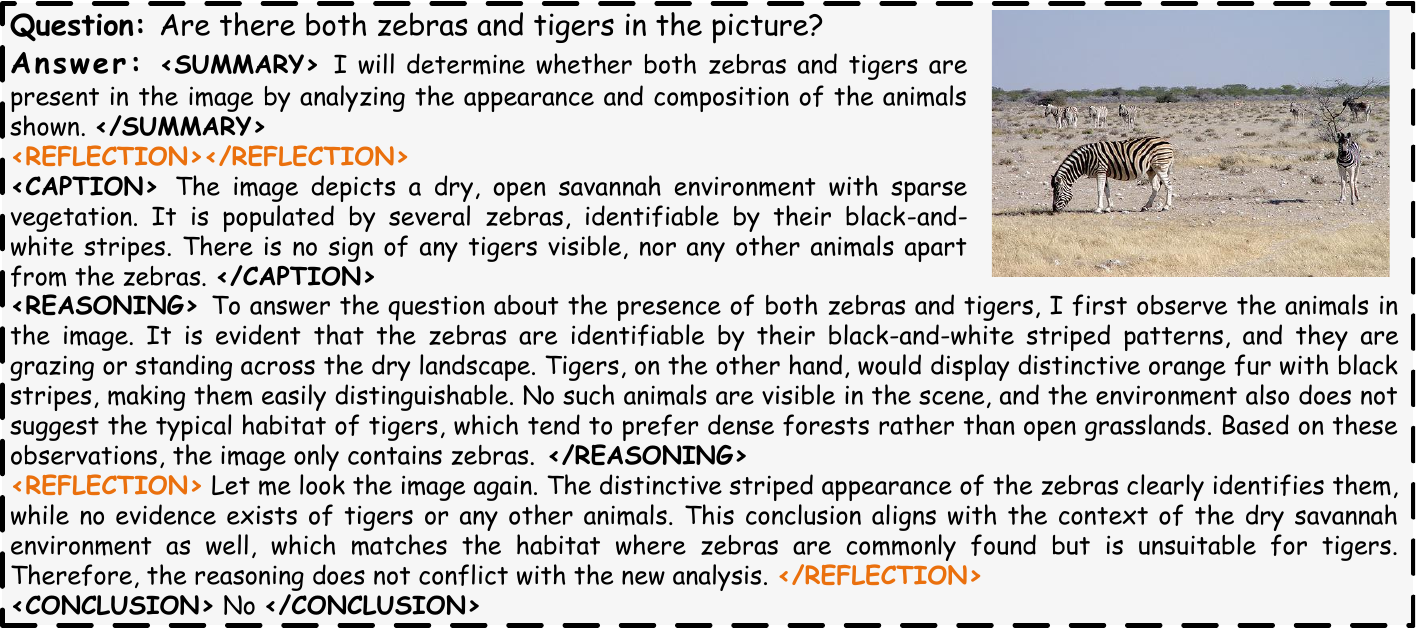}
    \caption{Qwen-Zero spontaneously develops multiple reflection awareness during BRPO training, including visual-only reflections without text, as shown in the first reflection in the example.}
    \label{fig:reflection_exp}
\end{figure*}

    $\bullet$ Some additional reflection processes produced by the model during BRPO training are empty in content, suggesting the model spontaneously develops a "Let me look at the image again" consciousness but does not always need to generate reflection text. As shown in Figure~\ref{fig:reflection_exp}.


Furthermore, we extract 40k questions from the LLaVA-CoT-100k dataset and use Qwen-Zero to generate responses for them to distill Qwen-Zero's reasoning and reflection capabilities. To ensure the correctness of model-generated results, we use both models and human verification, details of which can be found in Appendix~\ref{appendix:data validation}, to correct errors in Qwen-Zero's responses, creating the Qwen-Zero-40k dataset. Qwen-Zero-40k is used for supervised fine-tuning (SFT) of Qwen2.5-VL-7B-Instruct. Specifically, to force the model to re-attention visual information during generation, we propose and incorporate visual re-attention during both SFT and inference processes, with details provided in~\ref{Visual Re-attention}.

\subsection{Visual Re-attention}
\label{Visual Re-attention}
We first provide theoretical insights showing that hallucination in VLRMs relates to sequence length, with all theorem proofs available in Appendix~\ref{appendix:Detailed Proof}. Consistent with existing research~\cite{favero2024multi,zhou2023analyzing,huang2024opera,li2025hidden}, we assume that one key reason for VLM hallucination is visual information neglect, and provide proof based on this point.

Assume the number of text prompt tokens is $L_x$, image prompt tokens (visual tokens) is $L_c$, decoded sequence tokens is $L_y$, total sequence length is $L_{\text{total}} = L_x + L_c + L_y$. Model's output probability is:
\begin{equation}
p(y|x,c)=\prod_{t=1}^{L_y} p(y_t \mid y_{<t}, x, c),
\end{equation}
where $y_{<t} = [y_1,\dots,y_{t-1}]$. The visual context proportion in the sequence is $r = \frac{L_c}{L_{\text{total}}}$. Since we analyze overall visual attention rather than discussing attention differences of individual tokens, in the proof, we assume each token contributes roughly equal "information" given the context (text, visual, and generated tokens).

\textbf{Theorem 3.1} Visual attention decreases as generated content length increases during generation.

Mutual information measures the dependency between generated sequence $y$ and image prompt $c$ (given text prompt $x$), i.e., $I(y;c \mid x)=H(y\mid x)-H(y\mid x,c)$, where $H(\cdot)$ is entropy. Assuming similar entropy contribution from each token, we can approximate $I(y;c \mid x) \lesssim \frac{L_c}{L_{\text{total}}} \, H(y\mid x,c)$. Therefore, as $L_y$ increases during autoregressive generation, causing $L_{\text{total}}$ to increase, the ratio $r=\frac{L_c}{L_{\text{total}}}$ decreases, leading to decreased model attention (mutual information) to image prompt $c$.

Next, we formally prove that increasing visual token proportion can enhance visual attention. Different from~\cite{LiuZC24}, instead of adjusting visual token attention weights, we prove that introducing more visual tokens can enhance visual attention.

\textbf{Theorem 3.2} Repeated introduction of visual tokens increases visual attention.

If during generation, visual tokens are copied or partially routed and inserted into the generation sequence at one or multiple positions. Let $k>0$ be the number of copied inserted tokens, then the updated visual tokens number is $L_c' = L_c + k$, updated total sequence length is $L_{\text{total}}' = L_{\text{total}} + k$. The updated visual tokens ratio becomes $r' = \frac{L_c'}{L_{\text{total}}'} = \frac{L_c + k}{L_{\text{total}} + k}$. When input text prompt length $L_x>0$, we have $r' > r$. Therefore, the updated mutual information has upper bound $I(y;c\mid x) \lesssim r'\cdot H(y\mid x,c)$ higher than the pre-update upper bound, meaning increased visual attention during generation.

Based on the above analysis, during SFT and model inference, we propose two parallel methods to reintroduce visual tokens, driving model re-attention visuals. The two methods produce two variants of Qwen-LA: Qwen-LA-COPY and Qwen-LA-ROUTE.

\textbf{Visual Token COPY (VTC)}. When the model produces a reflection process (start with <REFLECTION> token), simply copy the input image's visual tokens completely to the beginning of the reflection process (before <REFLECTION> token). That is, the reflection process begins generating after the copied visual tokens.

\textbf{Visual Token ROUTE (VTR)}. When the model produces a reflection process (start with <REFLECTION> token), calculate previously generated tokens' attention to visual tokens, and route visual tokens with higher weights to the beginning of reflection process. Routing can be formalized as first calculating attention for all visual tokens:
\begin{equation}
\text{attn}_j = \frac{1}{L_y}\sum_{i=1}^{L_y}\text{attn}_{i,j}, \quad j \in \{1,2,\ldots,L_c\},
\end{equation}
where $\text{attn}_{i,j}$ represents attention weight of the $i$-th generated token to the $j$-th visual token, with $L_y$ here representing the decoded sequence length up to <REFLECTION> token. Then, rank visual tokens based on attention scores and select top $m\%$ visual tokens with highest attention for routing. 

\section{Experiment}
In this section, we evaluate Qwen-LA-COPY and Qwen-LA-ROUTE's accuracy metrics on visual QA datasets and hallucination metrics on hallucination datasets. Our work aims to address the following questions: 
\textbf{RQ1}: How is the accuracy and hallucination performance of Qwen-LA?
\textbf{RQ2}: What are the inference overheads of Qwen-LA-COPY and Qwen-LA-ROUTE? \textbf{RQ3}: How do BRPO, Visual Token COPY, and Visual Token ROUTE help improve Qwen-LA's performance?

\begin{table}[t]
\centering
\footnotesize
\caption{Accuracy (\%) of \textcolor{blue}{VLMs} and \textcolor{orange}{VLRMs} on visual QA datasets. \textbf{Bold} and \underline{underline} represent first and second place, respectively. 27B(4.5B) and 16B(3B) represent the total parameters of MoE~\cite{shazeer2017outrageously} architecture and the activated parameters (in parentheses).}
\label{tab:accuracy}
\resizebox{\textwidth}{!}{%
\begin{tabular}{l|c|ccccc}
\toprule
\textbf{Model} & \textbf{Parameters} & \textbf{MMMU} $\uparrow$ & \textbf{MMMU-Pro} $\uparrow$ & \textbf{MMBench} $\uparrow$ & \textbf{MMStar} $\uparrow$ & \textbf{MathVision} $\uparrow$ \\ \midrule
\textcolor{blue}{\textbf{Qwen2.5-VL-7B-Instruct}}~\cite{bai2025qwen2} & 7B & 58.6 & 41.0 & 82.6 & 63.9 & 25.1 \\
\textcolor{blue}{\textbf{Llama-3.2-11B-Vision-Instruct}}~\cite{meta2024llama32} & 11B & 50.7 & 33.0 & 64.9 & 46.6 & 12.4 \\
\textcolor{blue}{\textbf{MiniCPM-o-2.6}}~\cite{yao2024minicpm} & 8B & 50.4 & 30.0 & 78.0 & 57.5 & 23.1 \\
\textcolor{blue}{\textbf{InternVL2.5-8B}}~\cite{chen2024internvl} & 8B & 56.0 & 34.3 & 79.4 & 61.5 & 19.7 \\
\textcolor{blue}{\textbf{DeepSeek-VL2}}~\cite{wu2024deepseekvl2mixtureofexpertsvisionlanguagemodels} & 27B(4.5B) & 54.0 & 32.4 & 81.2 & 61.9 & 19.3 \\
\textcolor{blue}{\textbf{Gemma3-12B-IT}}~\cite{gemma_2025} & 12B & 50.3 & 29.1 & 79.4 & 58.5 & \textbf{28.6} \\ \midrule
\textcolor{orange}{\textbf{LLaVA-CoT}}~\cite{xu2411llava} & 11B & 51.2 & 31.7 & 73.8 & 57.8 & 15.6 \\
\textcolor{orange}{\textbf{InternVL2.5-8B-MPO}}~\cite{wang2024mpo} & 8B & 54.9 & 34.1 & 73.8 & \underline{65.7} & 18.0 \\
\textcolor{orange}{\textbf{Kimi-VL-Think}}~\cite{team2025kimi} & 16B(3B) & 57.0 & 35.4 & \textbf{83.1} & 61.3 & 21.4 \\
\textcolor{orange}{\textbf{Vision-R1}}~\cite{huang2025vision} & 7B & 56.2 & 36.1 & 81.5 & 61.4 & 25.5 \\ \midrule
\textbf{Qwen-LA-COPY (ours)} & 7B & \textbf{60.3} & \textbf{41.7} & 82.7 & \textbf{65.9} & \underline{26.4} \\
\textbf{Qwen-LA-ROUTE (ours)} & 7B & \underline{59.1} & \underline{41.3} & \underline{82.8} & 64.6 & 25.8 \\ \bottomrule
\end{tabular}%
}
\end{table}
\subsection{Experimental Setup}
\textbf{Datasets and metrics}. We select five widely used and challenging multimodal QA datasets for accuracy metric evaluation: MMMU~\cite{yue2024mmmu}, MMMU-Pro~\cite{yue2024mmmupro}, MMBench-V1.1-En~\cite{Liu2023MMBenchIY}, MMStar~\cite{Chen2024AreWO}, and MathVision~\cite{Wang2024MeasuringMM}. For hallucination evaluation, we first use MSCOCO~\cite{lin2014microsoft}, a comprehensive dataset used for image recognition, segmentation, and captioning. 5,000 unique images are selected from the COCO 2014 training dataset to evaluate hallucination performance~\cite{zhou2023analyzing}. The hallucination metrics evaluated on MSCOCO include: CHAIR~\cite{Rohrbach2018ObjectHI}, with two variants, $\text{CHAIR}\text{i}=\frac{\#\ \text{hallucinated objects}}{\#\ \text{generated objects}}$ calculating the proportion of hallucinated objects in the entire description, and $\text{CHAIR}\text{s}=\frac{\#\ \text{hallucinated captions}}{\#\ \text{generated captions}}$ evaluating the proportion of descriptions containing at least one object hallucination. Considering that sequence length significantly affects CHAIR values~\cite{Li2023EvaluatingOH}, we only retain the model's final conclusions for fair evaluation. POPE~\cite{Li2023EvaluatingOH} adopts a question-answering format to prompt the model, such as "Is there an <object> in the image?", to determine whether the model can correctly identify if specific objects are present in a given image. Additionally, MMHAL BENCH~\cite{Sun2023AligningLM} emphasizes logical reasoning and complex visual understanding, thus providing rigorous testing of hallucination mitigation in challenging scenarios. Model responses are evaluated using GPT-4o to align with ground-truth answers. MME~\cite{Fu2023MMEAC} covers 14 different visual-language capabilities, testing perception, reasoning, and knowledge integration through carefully curated image questions, thereby providing a holistic view of model performance.

\textbf{Baselines}. For visual QA tasks: We select Qwen2.5-VL-Instruct~\cite{bai2025qwen2}, Llama-3.2-11B-Vision-Instruct~\cite{meta2024llama32}, MiniCPM-o-2.6~\cite{yao2024minicpm}, InternVL2.5-8B~\cite{chen2024internvl}, DeepSeek-VL2~\cite{wu2024deepseekvl2mixtureofexpertsvisionlanguagemodels}, and Gemma3-12B-IT~\cite{gemma_2025} as VLM baselines. We select LLaVA-CoT~\cite{xu2411llava}, InternVL2.5-8B-MPO~\cite{wang2024mpo}, Kimi-VL-Think~\cite{team2025kimi}, and Vision-R1~\cite{huang2025vision} as VLRM baselines. For hallucination tasks, we select: M3ID~\cite{favero2024multi}, a training-free method that enhances the importance of visual prompts relative to language priors to improve visual grounding and reduce hallucination. Greedy Decode~\cite{zhou2023analyzing} abandons sampling strategies and aims to make the model output the most certain tokens. Chain-of-Thought (CoT)~\cite{Wei2022ChainOT}, where we use prompts to make the model generate answers after step-by-step description. OPERA~\cite{huang2024opera} introduces penalty terms for model logic in beam-search decoding to mitigate over-trust issues. VISTA~\cite{li2025hidden} strengthens visual information in activation space and utilizes early layer activations to promote meaningful semantic decoding. ICoT~\cite{interleaved2025} adopts multimodal prompting to enhance visual information. PAI~\cite{LiuZC24} adaptively adjusts and amplifies attention weights assigned to image tokens, thereby highlighting visual elements. We also evaluate Qwen2.5-VL-Instruct as base model.

\textbf{Implementation Details}. We set the ideal single reflection average length $\lambda$ in BPRO to 100 and the percentage $m$ in Visual Token ROUTE to 50. Other training process parameter settings details are provided in Appendix~\ref{appendix:Training Parameter Details}.
\subsection{Main Results}
We evaluate Qwen-LA's accuracy performance on visual QA datasets, as shown in Table~\ref{tab:accuracy}. Compared to VLMs, especially the base model Qwen2.5-VL-7B-Instruct, Qwen-LA achieves performance improvements across all tasks. When compared to VLRMs, despite using only 40k fine-tuning data, Qwen-LA, particularly Qwen-LA-COPY, achieves leading performance on most tasks.

\begin{table}[t]
\centering
\footnotesize
\caption{Hallucination Metrics of  different models/methods. We report the average F1 score of POPE.}
\label{tab:hallucination}
\begin{tabular}{l|ccccc}
\toprule
\textbf{Model / Method} & \textbf{CHAIR}$_\text{i}$ $\downarrow$ & \textbf{CHAIR}$_\text{s}$ $\downarrow$ & \textbf{POPE} $\uparrow$ & \textbf{MMHAL BENCH} $\uparrow$ & \textbf{MME} $\uparrow$ \\ \midrule
\textbf{M3ID}~\cite{favero2024multi} & 5.9 & 13.8 & 76.0 & - & - \\
\textbf{Greedy Decode}~\cite{zhou2023analyzing} & 9.1 & 36.4 & 88.2 & 3.62 & 2310.2 \\
\textbf{CoT}~\cite{Wei2022ChainOT} & 7.9 & 40.8 & 88.5 & 3.71 & 2314.0 \\
\textbf{OPERA}~\cite{huang2024opera} & 12.4 & 45.2 & 85.8 & 2.33 & 1515.4 \\
\textbf{VISTA}~\cite{li2025hidden} & 6.3 & 17.4 & 85.9 & 2.95 & 1738.5 \\
\textbf{ICoT}~\cite{interleaved2025} & 8.3 & 31.2 & 88.6 & 3.38 & 2227.5 \\
\textbf{PAI}~\cite{LiuZC24} & 6.8 & 22.3 & 85.9 & 2.41 & 1644.0 \\
\textbf{Qwen2.5-VL-Instruct}~\cite{bai2025qwen2} & 9.4 & 37.1 & \underline{88.7} & 3.68 & 2309.4 \\ \midrule
\textbf{Qwen-LA-COPY (ours)} & \textbf{3.7} & \textbf{9.8} & \textbf{90.2} & \textbf{3.82} & \textbf{2330.8} \\
\textbf{Qwen-LA-ROUTE (ours)} & \underline{5.6} & \underline{11.2} & 88.5 & \underline{3.73} & \underline{2322.6} \\ \bottomrule
\end{tabular}
\end{table}

We evaluate Qwen-LA's hallucination metrics on hallucination datasets, as shown in Table~\ref{tab:hallucination}. In particular, ICoT~\cite{interleaved2025} and PAI~\cite{LiuZC24} form a contrast with our VTC and VTR. ICoT adds visual-text prompts composed of local image features before each reasoning paragraph. PAI dynamically adjusts the attention weights of visual tokens without supplementing additional visual tokens. The results demonstrate that Qwen-LA effectively mitigates hallucination. Qwen-LA-COPY performs slightly better than Qwen-LA-ROUTE, which routes partial tokens, due to its complete copying of visual tokens during reflection. However, the latter still achieves competitive performance.

\subsection{Inference Time Overhead}
\begin{table}[t]
\begin{minipage}{0.43\textwidth}
\centering
\footnotesize
\caption{Comparison of accuracy, generation length, and inference time for different models on MMMU.}
\label{tab:time_overhead}
\scalebox{0.750}{
\begin{tabular}{l|ccc}
\toprule
\textbf{Models} & \textbf{ACC(\%)} & \textbf{Length} & \textbf{time(s)} \\ \midrule
\textbf{Qwen2.5-VL-Instruct} & 58.6 & 268.5 & \textbf{8.68} \\
\textbf{Kimi-VL-Think} & 57.0 & \underline{1572.5} & 99.08 \\
\textbf{Vision-R1} & 56.2 & 358.9 & 120.37 \\ \midrule
\textbf{Qwen-LA-COPY} & \textbf{60.3} & \textbf{1811.4} & 22.33 \\
\textbf{Qwen-LA-ROUTE} & \underline{59.1} & 1425.8 & \underline{18.29} \\ \bottomrule
\end{tabular}
}
\end{minipage}
\hfill
\begin{minipage}{0.55\textwidth}
\centering
\footnotesize
\caption{Ablation study of different components in Qwen-LA on QA and hallucination datasets.}
\label{tab:ablation}
\scalebox{0.750}{
\begin{tabular}{l|cccc}
\toprule
\textbf{Method} & \textbf{MMMU} & \textbf{MMStar} & \textbf{CHAIR}$_\text{i}$ & \textbf{MME} \\ \midrule
\textbf{Qwen2.5-VL-Instruct} & 58.6 & 63.9 & 9.4 & 2309.4 \\ \midrule
\textbf{w/o BRPO (w/ VTC)} & 57.2 & 62.2 & 6.4 & 2312.2 \\
\textbf{w/o BRPO (w/ VTR)} & 57.0 & 61.3 & 6.2 & 2311.9 \\
\textbf{w/ BRPO (w/o VTC / VTR)} & 58.8 & 64.2 & 8.7 & 2308.5 \\ \midrule
\textbf{Qwen-LA-COPY} & \textbf{60.3} & \textbf{65.9} & \textbf{3.7} & \textbf{2330.8} \\
\textbf{Qwen-LA-ROUTE} & \underline{59.1} & \underline{64.6} & \underline{5.6} & \underline{2322.6} \\ \bottomrule
\end{tabular}
}
\end{minipage}
\end{table}

We compare the accuracy, output sequence length, and inference time of different models on MMMU. Although different models use different tokenizers, we still use token count to measure sequence length considering that inference overhead correlates with the number of decoded tokens. As shown in Table~\ref{tab:time_overhead}, compared to other reasoning models like Kimi-VL-Think and Vision-R1, Qwen-LA has longer sequence length but shorter inference time. Even when compared to non-reasoning models (such as Qwen2.5-VL-Instruct), Qwen-LA achieves leading accuracy with only increasing reasoning time within an acceptable range. Moreover, setting the hyperparameter $m$ to a lower value can further reduce Qwen-LA-ROUTE's time overhead, as discussed in Table~\ref{tab:m_values}. Although Qwen-LA-COPY has longer inference time, it achieves higher accuracy. We believe these increases in inference time are acceptable in scenarios where model accuracy needs to be enhanced.

\subsection{Ablation Study}

Table~\ref{tab:ablation} shows the ablation experiments for different components of Qwen-LA. "w/o BRPO" indicates not using BRPO, but directly prompting Qwen2.5-VL-Instruct to generate reasoning and reflection, using either VTC or VTR during reflection. "w/ BRPO" indicates using BRPO but without visual re-attention methods. It can be observed that "w/o BRPO" shows performance degradation due to unstable outputs without BRPO and SFT. Using BRPO without VTC/VTR (i.e., text-only reflection) also fails to achieve optimal performance. Qwen-LA-COPY, which uses both BRPO and VTC, achieves the best performance, while using VTR also brings performance improvements.

\textbf{Visual re-attention enhanced visual token attention weights}. We measure the average attention to visual tokens for Qwen-LA-COPY and Qwen-LA-ROUTE on 500 randomly selected samples from the MSCOCO dataset, as shown in Figure~\ref{fig:attn_vtc_vtr}. We calculate the mean attention weights of all visual tokens across all layers. Compared to models without visual re-attention (Figure~\ref{fig:org_attn}), both VTC and VTR enhance the attention weights of visual tokens during the generation process. This may be beneficial for mitigating hallucinations in VLMs, as the occurrence of hallucinations might be related to reduced visual attention~\cite{favero2024multi,zhou2023analyzing,huang2024opera,li2025hidden}.

\begin{figure*}[t]
\centering
\includegraphics[width=\textwidth]{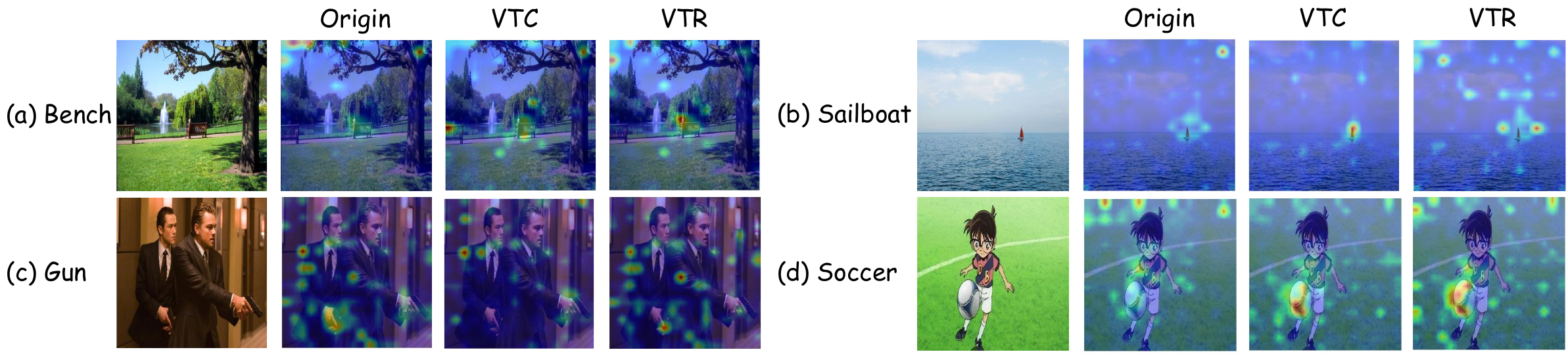}
\caption{Attention visualization of VTC and VTR.}
\label{fig:visualization}
\vspace{-6pt}
\end{figure*}

\begin{table}[t]
\centering
\footnotesize
\caption{Results of different $m$ for Qwen-LA-ROUTE on visual QA and hallucination datasets.}
\label{tab:m_values}
\begin{tabular}{l|c|c|c|c|c|c|c|c}
\toprule
\multirow{2}{*}{\textbf{$m (\%)$}} & \multicolumn{2}{c|}{\textbf{MMMU}} & \multicolumn{2}{c|}{\textbf{MMStar}} & \multicolumn{2}{c|}{\textbf{POPE}} & \multicolumn{2}{c}{\textbf{MME}} \\
& \textbf{Acc(\%)} & \textbf{time(s)} & \textbf{Acc(\%)} & \textbf{time(s)} & \textbf{score} & \textbf{time(s)} & \textbf{score} & \textbf{time(s)} \\ \midrule
\textbf{0}        & 58.8& \textbf{17.52}& 64.2& \textbf{17.80}& 88.4& \textbf{15.64}& 2308.5& \textbf{14.02}\\
\textbf{25}       & 55.6& \underline{17.90}& 62.1& \underline{18.36}& 86.2& \underline{16.53}& 2290.2& \underline{16.38}\\
\textbf{50}       & 59.1& 18.29& \underline{64.6}& 19.07& 88.5& 16.66& \underline{2322.6}& 16.94\\
\textbf{75}       & \underline{59.5}& 20.31& \underline{64.6}& 19.52& \underline{88.7}& 16.98& 2322.3& 17.27\\
\textbf{100}      & \textbf{60.3}& 22.33& \textbf{65.9}& 19.89& \textbf{90.2}& 17.25& \textbf{2330.8}& 17.60\\ \bottomrule
\end{tabular}
\end{table}

\textbf{Visualization of VTC and VTR.} We visualize attention heatmaps of visual re-attention, as shown in Figure~\ref{fig:visualization}. We overlaid attention heatmaps from all layers of Qwen-LA  on the images, as different layers may extract different information from the image. Only heatmaps when generating key tokens (text shown on the left side of images in Figure~\ref{fig:visualization}) are printed. We observe that compared to the first view of the image (Origin), VTC focus on other content in the image, typically supplementing details that are not attended to during the first viewing. 
Notably, VTC observe two benches in example (a) while Origin only notice one. VTR, by routing visual tokens with high attention weights, enhance the image patches that are attended to in the first view. For example, VTR enhance attention to the left man's hands in example (c). Although VTC and VTR enhance visual information in different ways, both approaches led to performance improvements and reduce hallucination.


\textbf{VTR Parameter $m\%$.} VTR routes the top $m\%$ visual tokens to the reflection process. The hyperparameter $m$ may affects model performance and inference time. Table~\ref{tab:m_values} shows the model's accuracy, hallucination metrics, and inference time overhead under different $m$. Larger $m$ lead to higher accuracy and lower hallucination, while increasing inference time. Unfortunately, small $m$, like 25, reduce model accuracy. Therefore, we encourage selecting $m$ based on the real-time or accuracy requirements of the task to balance performance and time costs.

\section{Conclusion}
In this paper, we propose Qwen-LookAgain (Qwen-LA), a novel VLRM that introduces a vision-text reflection process with visual re-attention during reasoning. We propose BRPO to guide the model in spontaneously generating a multi-stage reflection process. Furthermore, we introduce Visual Token COPY and Visual Token ROUTE in both training and inference phases to force the model to re-attend to visual information. Experiments on multiple visual QA datasets and hallucination metrics show that Qwen-LA improves accuracy and reduces hallucinations.

\section{Limitations}
\label{sec:limitation}
In our formal analysis of the relationship between VLM hallucination and generation length, we assume that one important cause of VLM hallucination is the neglect of visual information. While this is a common perspective and analytical approach~\cite{favero2024multi,zhou2023analyzing,huang2024opera,li2025hidden}, it may limit the scope of the proof. Additionally, both proposed VTC and VTR increase the inference overhead of VLRM, which may not be acceptable in scenarios requiring instant response. We encourage the use of our methods in scenarios where high accuracy is required. Finally, although we only propose our strategies based on Qwen2.5-VL-Instruct due to resource limitations, we declare that our methods are extensible to other VLMs, and we encourage the community to experiment with stronger base models.

\bibliographystyle{plain}  
\bibliography{references}   

\begin{thebibliography}{10}

\bibitem{aggarwal2025l1}
Pranjal Aggarwal and Sean Welleck.
\newblock L1: Controlling how long a reasoning model thinks with reinforcement learning.
\newblock {\em arXiv preprint arXiv:2503.04697}, 2025.

\bibitem{bai2025qwen2}
Shuai Bai, Keqin Chen, Xuejing Liu, Jialin Wang, Wenbin Ge, Sibo Song, Kai Dang, Peng Wang, Shijie Wang, Jun Tang, et~al.
\newblock Qwen2. 5-vl technical report.
\newblock {\em arXiv preprint arXiv:2502.13923}, 2025.

\bibitem{chen2024step}
Guoxin Chen, Minpeng Liao, Chengxi Li, and Kai Fan.
\newblock Step-level value preference optimization for mathematical reasoning.
\newblock {\em arXiv preprint arXiv:2406.10858}, 2024.

\bibitem{Chen2024AreWO}
Lin Chen, Jinsong Li, Xiao wen Dong, Pan Zhang, Yuhang Zang, Zehui Chen, Haodong Duan, Jiaqi Wang, Yu~Qiao, Dahua Lin, and Feng Zhao.
\newblock Are we on the right way for evaluating large vision-language models?
\newblock {\em ArXiv}, abs/2403.20330, 2024.

\bibitem{chen2025towards}
Qiguang Chen, Libo Qin, Jinhao Liu, Dengyun Peng, Jiannan Guan, Peng Wang, Mengkang Hu, Yuhang Zhou, Te~Gao, and Wanxiang Che.
\newblock Towards reasoning era: A survey of long chain-of-thought for reasoning large language models.
\newblock {\em arXiv preprint arXiv:2503.09567}, 2025.

\bibitem{chen2024internvl}
Zhe Chen, Jiannan Wu, Wenhai Wang, Weijie Su, Guo Chen, Sen Xing, Muyan Zhong, Qinglong Zhang, Xizhou Zhu, Lewei Lu, et~al.
\newblock Internvl: Scaling up vision foundation models and aligning for generic visual-linguistic tasks.
\newblock In {\em Proceedings of the IEEE/CVF Conference on Computer Vision and Pattern Recognition}, pages 24185--24198, 2024.

\bibitem{chu2025domaino1s}
Xu~Chu, Zhijie Tan, Hanlin Xue, Guanyu Wang, Tong Mo, and Weiping Li.
\newblock Domaino1s: Guiding llm reasoning for explainable answers in high-stakes domains.
\newblock {\em arXiv preprint arXiv:2501.14431}, 2025.

\bibitem{favero2024multi}
Alessandro Favero, Luca Zancato, Matthew Trager, Siddharth Choudhary, Pramuditha Perera, Alessandro Achille, Ashwin Swaminathan, and Stefano Soatto.
\newblock Multi-modal hallucination control by visual information grounding.
\newblock In {\em Proceedings of the IEEE/CVF Conference on Computer Vision and Pattern Recognition}, pages 14303--14312, 2024.

\bibitem{Fu2023MMEAC}
Chaoyou Fu, Peixian Chen, Yunhang Shen, Yulei Qin, Mengdan Zhang, Xu~Lin, Zhenyu Qiu, Wei Lin, Jinrui Yang, Xiawu Zheng, Ke~Li, Xing Sun, and Rongrong Ji.
\newblock Mme: A comprehensive evaluation benchmark for multimodal large language models.
\newblock {\em ArXiv}, abs/2306.13394, 2023.

\bibitem{interleaved2025}
Jun Gao, Yongqi Li, Ziqiang Cao, and Wenjie Li.
\newblock Interleaved-modal chain-of-thought.
\newblock In {\em Proceedings of the IEEE/CVF Conference on Computer Vision and Pattern Recognition (CVPR)}, 2025.

\bibitem{guo2025deepseek}
Daya Guo, Dejian Yang, Haowei Zhang, Junxiao Song, Ruoyu Zhang, Runxin Xu, Qihao Zhu, Shirong Ma, Peiyi Wang, Xiao Bi, et~al.
\newblock Deepseek-r1: Incentivizing reasoning capability in llms via reinforcement learning.
\newblock {\em arXiv preprint arXiv:2501.12948}, 2025.

\bibitem{gupta2024polymath}
Himanshu Gupta, Shreyas Verma, Ujjwala Anantheswaran, Kevin Scaria, Mihir Parmar, Swaroop Mishra, and Chitta Baral.
\newblock Polymath: A challenging multi-modal mathematical reasoning benchmark.
\newblock {\em arXiv preprint arXiv:2410.14702}, 2024.

\bibitem{huang2024opera}
Qidong Huang, Xiaoyi Dong, Pan Zhang, Bin Wang, Conghui He, Jiaqi Wang, Dahua Lin, Weiming Zhang, and Nenghai Yu.
\newblock Opera: Alleviating hallucination in multi-modal large language models via over-trust penalty and retrospection-allocation.
\newblock In {\em Proceedings of the IEEE/CVF Conference on Computer Vision and Pattern Recognition}, pages 13418--13427, 2024.

\bibitem{huang2025vision}
Wenxuan Huang, Bohan Jia, Zijie Zhai, Shaosheng Cao, Zheyu Ye, Fei Zhao, Zhe Xu, Yao Hu, and Shaohui Lin.
\newblock Vision-r1: Incentivizing reasoning capability in multimodal large language models.
\newblock {\em arXiv preprint arXiv:2503.06749}, 2025.

\bibitem{jaech2024openai}
Aaron Jaech, Adam Kalai, Adam Lerer, Adam Richardson, Ahmed El-Kishky, Aiden Low, Alec Helyar, Aleksander Madry, Alex Beutel, Alex Carney, et~al.
\newblock Openai o1 system card.
\newblock {\em arXiv preprint arXiv:2412.16720}, 2024.

\bibitem{Labrak2024BioMistralAC}
Yanis Labrak, Adrien Bazoge, Emmanuel Morin, Pierre-Antoine Gourraud, Mickael Rouvier, and Richard Dufour.
\newblock Biomistral: A collection of open-source pretrained large language models for medical domains.
\newblock In {\em Annual Meeting of the Association for Computational Linguistics}, 2024.

\bibitem{Li2023EvaluatingOH}
Yifan Li, Yifan Du, Kun Zhou, Jinpeng Wang, Wayne~Xin Zhao, and Ji~rong Wen.
\newblock Evaluating object hallucination in large vision-language models.
\newblock In {\em Conference on Empirical Methods in Natural Language Processing}, 2023.

\bibitem{li2025hidden}
Zhuowei Li, Haizhou Shi, Yunhe Gao, Di~Liu, Zhenting Wang, Yuxiao Chen, Ting Liu, Long Zhao, Hao Wang, and Dimitris~N Metaxas.
\newblock The hidden life of tokens: Reducing hallucination of large vision-language models via visual information steering.
\newblock {\em arXiv preprint arXiv:2502.03628}, 2025.

\bibitem{liang2024scemqa}
Zhenwen Liang, Kehan Guo, Gang Liu, Taicheng Guo, Yujun Zhou, Tianyu Yang, Jiajun Jiao, Renjie Pi, Jipeng Zhang, and Xiangliang Zhang.
\newblock Scemqa: A scientific college entrance level multimodal question answering benchmark.
\newblock {\em arXiv preprint arXiv:2402.05138}, 2024.

\bibitem{liao2025longperceptualthoughts}
Yuan-Hong Liao, Sven Elflein, Liu He, Laura Leal-Taix{\'e}, Yejin Choi, Sanja Fidler, and David Acuna.
\newblock Longperceptualthoughts: Distilling system-2 reasoning for system-1 perception.
\newblock {\em arXiv preprint arXiv:2504.15362}, 2025.

\bibitem{lin2014microsoft}
Tsung-Yi Lin, Michael Maire, Serge Belongie, James Hays, Pietro Perona, Deva Ramanan, Piotr Doll{\'a}r, and C~Lawrence Zitnick.
\newblock Microsoft coco: Common objects in context.
\newblock In {\em Computer vision--ECCV 2014: 13th European conference, zurich, Switzerland, September 6-12, 2014, proceedings, part v 13}, pages 740--755. Springer, 2014.

\bibitem{liu2024survey}
Hanchao Liu, Wenyuan Xue, Yifei Chen, Dapeng Chen, Xiutian Zhao, Ke~Wang, Liping Hou, Rongjun Li, and Wei Peng.
\newblock A survey on hallucination in large vision-language models.
\newblock {\em arXiv preprint arXiv:2402.00253}, 2024.

\bibitem{LiuZC24}
Shi Liu, Kecheng Zheng, and Wei Chen.
\newblock Paying more attention to image: A training-free method for alleviating hallucination in lvlms.
\newblock In {\em European Conference on Computer Vision}, pages 125--140. Springer, 2024.

\bibitem{Liu2023MMBenchIY}
Yuanzhan Liu, Haodong Duan, Yuanhan Zhang, Bo~Li, Songyang Zhang, Wangbo Zhao, Yike Yuan, Jiaqi Wang, Conghui He, Ziwei Liu, Kai Chen, and Dahua Lin.
\newblock Mmbench: Is your multi-modal model an all-around player?
\newblock In {\em European Conference on Computer Vision}, 2023.

\bibitem{liu2025visual}
Ziyu Liu, Zeyi Sun, Yuhang Zang, Xiaoyi Dong, Yuhang Cao, Haodong Duan, Dahua Lin, and Jiaqi Wang.
\newblock Visual-rft: Visual reinforcement fine-tuning.
\newblock {\em arXiv preprint arXiv:2503.01785}, 2025.

\bibitem{lu2021inter}
Pan Lu, Ran Gong, Shibiao Jiang, Liang Qiu, Siyuan Huang, Xiaodan Liang, and Song-Chun Zhu.
\newblock Inter-gps: Interpretable geometry problem solving with formal language and symbolic reasoning.
\newblock In {\em The Joint Conference of the 59th Annual Meeting of the Association for Computational Linguistics and the 11th International Joint Conference on Natural Language Processing (ACL-IJCNLP 2021)}, 2021.

\bibitem{madaan2023self}
Aman Madaan, Niket Tandon, Prakhar Gupta, Skyler Hallinan, Luyu Gao, Sarah Wiegreffe, Uri Alon, Nouha Dziri, Shrimai Prabhumoye, Yiming Yang, et~al.
\newblock Self-refine: Iterative refinement with self-feedback.
\newblock {\em Advances in Neural Information Processing Systems}, 36:46534--46594, 2023.

\bibitem{meta2024llama32}
{Meta}.
\newblock Llama 3.2: Revolutionizing edge ai and vision with open, customizable models.
\newblock Technical report, Meta, 2024.

\bibitem{ouyang2022training}
Long Ouyang, Jeffrey Wu, Xu~Jiang, Diogo Almeida, Carroll Wainwright, Pamela Mishkin, Chong Zhang, Sandhini Agarwal, Katarina Slama, Alex Ray, et~al.
\newblock Training language models to follow instructions with human feedback.
\newblock {\em Advances in neural information processing systems}, 35:27730--27744, 2022.

\bibitem{qiao2024we}
Runqi Qiao, Qiuna Tan, Guanting Dong, Minhui Wu, Chong Sun, Xiaoshuai Song, Zhuoma GongQue, Shanglin Lei, Zhe Wei, Miaoxuan Zhang, et~al.
\newblock We-math: Does your large multimodal model achieve human-like mathematical reasoning?
\newblock {\em arXiv preprint arXiv:2407.01284}, 2024.

\bibitem{Rohrbach2018ObjectHI}
Anna Rohrbach, Lisa~Anne Hendricks, Kaylee Burns, Trevor Darrell, and Kate Saenko.
\newblock Object hallucination in image captioning.
\newblock In {\em Conference on Empirical Methods in Natural Language Processing}, 2018.

\bibitem{shao2024deepseekmath}
Zhihong Shao, Peiyi Wang, Qihao Zhu, Runxin Xu, Junxiao Song, Xiao Bi, Haowei Zhang, Mingchuan Zhang, YK~Li, Y~Wu, et~al.
\newblock Deepseekmath: Pushing the limits of mathematical reasoning in open language models.
\newblock {\em arXiv preprint arXiv:2402.03300}, 2024.

\bibitem{shazeer2017outrageously}
Noam Shazeer, Azalia Mirhoseini, Krzysztof Maziarz, Andy Davis, Quoc Le, Geoffrey Hinton, and Jeff Dean.
\newblock Outrageously large neural networks: The sparsely-gated mixture-of-experts layer.
\newblock {\em arXiv preprint arXiv:1701.06538}, 2017.

\bibitem{Sun2023AligningLM}
Zhiqing Sun, Sheng Shen, Shengcao Cao, Haotian Liu, Chunyuan Li, Yikang Shen, Chuang Gan, Liangyan Gui, Yu-Xiong Wang, Yiming Yang, Kurt Keutzer, and Trevor Darrell.
\newblock Aligning large multimodal models with factually augmented rlhf.
\newblock {\em ArXiv}, abs/2309.14525, 2023.

\bibitem{gemma_2025}
Gemma Team.
\newblock Gemma 3.
\newblock Technical report, Google DeepMind, 2025.

\bibitem{team2025kimi}
Kimi Team, Angang Du, Bohong Yin, Bowei Xing, Bowen Qu, Bowen Wang, Cheng Chen, Chenlin Zhang, Chenzhuang Du, Chu Wei, et~al.
\newblock Kimi-vl technical report.
\newblock {\em arXiv preprint arXiv:2504.07491}, 2025.

\bibitem{qwq32b}
Qwen Team.
\newblock Qwq-32b: Embracing the power of reinforcement learning, March 2025.

\bibitem{wang2024picture}
Jiayu Wang, Yifei Ming, Zhenmei Shi, Vibhav Vineet, Xin Wang, Sharon Li, and Neel Joshi.
\newblock Is a picture worth a thousand words? delving into spatial reasoning for vision language models.
\newblock {\em Advances in Neural Information Processing Systems}, 37:75392--75421, 2024.

\bibitem{wang2024measuring}
Ke~Wang, Junting Pan, Weikang Shi, Zimu Lu, Houxing Ren, Aojun Zhou, Mingjie Zhan, and Hongsheng Li.
\newblock Measuring multimodal mathematical reasoning with math-vision dataset.
\newblock {\em Advances in Neural Information Processing Systems}, 37:95095--95169, 2024.

\bibitem{Wang2024MeasuringMM}
Ke~Wang, Junting Pan, Weikang Shi, Zimu Lu, Houxing Ren, Aojun Zhou, Mingjie Zhan, and Hongsheng Li.
\newblock Measuring multimodal mathematical reasoning with math-vision dataset.
\newblock In {\em Neural Information Processing Systems}, 2024.

\bibitem{wang2024mpo}
Weiyun Wang, Zhe Chen, Wenhai Wang, Yue Cao, Yangzhou Liu, Zhangwei Gao, Jinguo Zhu, Xizhou Zhu, Lewei Lu, Yu~Qiao, and Jifeng Dai.
\newblock Enhancing the reasoning ability of multimodal large language models via mixed preference optimization.
\newblock {\em arXiv preprint arXiv:2411.10442}, 2024.

\bibitem{Wei2022ChainOT}
Jason Wei, Xuezhi Wang, Dale Schuurmans, Maarten Bosma, Ed~H. Chi, F.~Xia, Quoc Le, and Denny Zhou.
\newblock Chain of thought prompting elicits reasoning in large language models.
\newblock {\em ArXiv}, abs/2201.11903, 2022.

\bibitem{wu2025effectively}
Tong Wu, Chong Xiang, Jiachen~T Wang, and Prateek Mittal.
\newblock Effectively controlling reasoning models through thinking intervention.
\newblock {\em arXiv preprint arXiv:2503.24370}, 2025.

\bibitem{wu2024deepseekvl2mixtureofexpertsvisionlanguagemodels}
Zhiyu Wu, Xiaokang Chen, Zizheng Pan, Xingchao Liu, Wen Liu, Damai Dai, Huazuo Gao, Yiyang Ma, Chengyue Wu, Bingxuan Wang, Zhenda Xie, Yu~Wu, Kai Hu, Jiawei Wang, Yaofeng Sun, Yukun Li, Yishi Piao, Kang Guan, Aixin Liu, Xin Xie, Yuxiang You, Kai Dong, Xingkai Yu, Haowei Zhang, Liang Zhao, Yisong Wang, and Chong Ruan.
\newblock Deepseek-vl2: Mixture-of-experts vision-language models for advanced multimodal understanding, 2024.

\bibitem{xie2024monte}
Yuxi Xie, Anirudh Goyal, Wenyue Zheng, Min-Yen Kan, Timothy~P Lillicrap, Kenji Kawaguchi, and Michael Shieh.
\newblock Monte carlo tree search boosts reasoning via iterative preference learning.
\newblock In {\em NeurIPS 2024 Workshop on System 2 Reasoning (Sys2-Reasoning)}, 2024.

\bibitem{xu2411llava}
Guowei Xu, Peng Jin, Li~Hao, Yibing Song, Lichao Sun, and Li~Yuan.
\newblock Llava-cot: Let vision language models reason step-by-step, 2024.
\newblock {\em URL https://arxiv. org/abs/2411.10440}, 2024.

\bibitem{yao2024minicpm}
Yuan Yao, Tianyu Yu, Ao~Zhang, Chongyi Wang, Junbo Cui, Hongji Zhu, Tianchi Cai, Haoyu Li, Weilin Zhao, Zhihui He, et~al.
\newblock Minicpm-v: A gpt-4v level mllm on your phone.
\newblock {\em arXiv preprint arXiv:2408.01800}, 2024.

\bibitem{yu2025introducing}
Runpeng Yu, Xinyin Ma, and Xinchao Wang.
\newblock Introducing visual perception token into multimodal large language model.
\newblock {\em arXiv preprint arXiv:2502.17425}, 2025.

\bibitem{yuan2024advancing}
Lifan Yuan, Ganqu Cui, Hanbin Wang, Ning Ding, Xingyao Wang, Jia Deng, Boji Shan, Huimin Chen, Ruobing Xie, Yankai Lin, et~al.
\newblock Advancing llm reasoning generalists with preference trees.
\newblock In {\em ICML 2024 Workshop on AI for Math}, 2024.

\bibitem{yue2024mmmu}
Xiang Yue, Yuansheng Ni, Kai Zhang, Tianyu Zheng, Ruoqi Liu, Ge~Zhang, Samuel Stevens, Dongfu Jiang, Weiming Ren, Yuxuan Sun, et~al.
\newblock Mmmu: A massive multi-discipline multimodal understanding and reasoning benchmark for expert agi.
\newblock In {\em Proceedings of the IEEE/CVF Conference on Computer Vision and Pattern Recognition}, pages 9556--9567, 2024.

\bibitem{yue2024mmmupro}
Xiang Yue, Tianyu Zheng, Yuansheng Ni, Yubo Wang, Kai Zhang, Shengbang Tong, Yuxuan Sun, Botao Yu, Ge~Zhang, Huan Sun, et~al.
\newblock Mmmu-pro: A more robust multi-discipline multimodal understanding benchmark.
\newblock {\em arXiv preprint arXiv:2409.02813}, 2024.

\bibitem{zhang-etal-2024-self-contrast}
Wenqi Zhang, Yongliang Shen, Linjuan Wu, Qiuying Peng, Jun Wang, Yueting Zhuang, and Weiming Lu.
\newblock Self-contrast: Better reflection through inconsistent solving perspectives.
\newblock In Lun-Wei Ku, Andre Martins, and Vivek Srikumar, editors, {\em Proceedings of the 62nd Annual Meeting of the Association for Computational Linguistics (Volume 1: Long Papers)}, pages 3602--3622, Bangkok, Thailand, August 2024. Association for Computational Linguistics.

\bibitem{zhou2023analyzing}
Yiyang Zhou, Chenhang Cui, Jaehong Yoon, Linjun Zhang, Zhun Deng, Chelsea Finn, Mohit Bansal, and Huaxiu Yao.
\newblock Analyzing and mitigating object hallucination in large vision-language models.
\newblock {\em arXiv preprint arXiv:2310.00754}, 2023.

\end{thebibliography}







\appendix
\section{BRPO Experimental Details}
\label{appendix:BRPO Experimental Details}
Before BRPO training, we first initialize with a cold start of Qwen2.5-VL-Instruct, with training details available in Appendix~\ref{appendix:Training Parameter Details}.

We train BRPO with full parameters on 8 NVIDIA A800-SXM4 80G GPUs for ~48 hours. The max length is set to 8192, with max completion length of 1024. We use bfloat16 precision for training, with a learning rate of 1.0e-6, $\beta$ value of 0.1, batch size of 16, and gradient accumulation steps of 2. We employ the AdamW optimizer with beta coefficients of 0.9 and 0.95, and an epsilon value of 1e-08. The training process employs DeepSpeed Zero-3 optimization strategy with a 0.01 warmup ratio. Each inference generates 8 candidate answers with a temperature coefficient of 1.0. We use the vllm engine with a maximum model length of 1024. For dataset images, we set a resolution limit of 640 $\times$ 480 and resize images exceeding this limit. This is merely to avoid memory explosion, as the number of visual tokens encoded by Qwen2.5-VL-Instruct correlates positively with input image resolution. We encourage removing this limitation when computational resources are sufficient to allow the model to adapt to images of different resolutions.

\begin{figure*}[ht]
    \centering
    \begin{minipage}{0.48\textwidth}
        \centering
        \includegraphics[width=\textwidth]{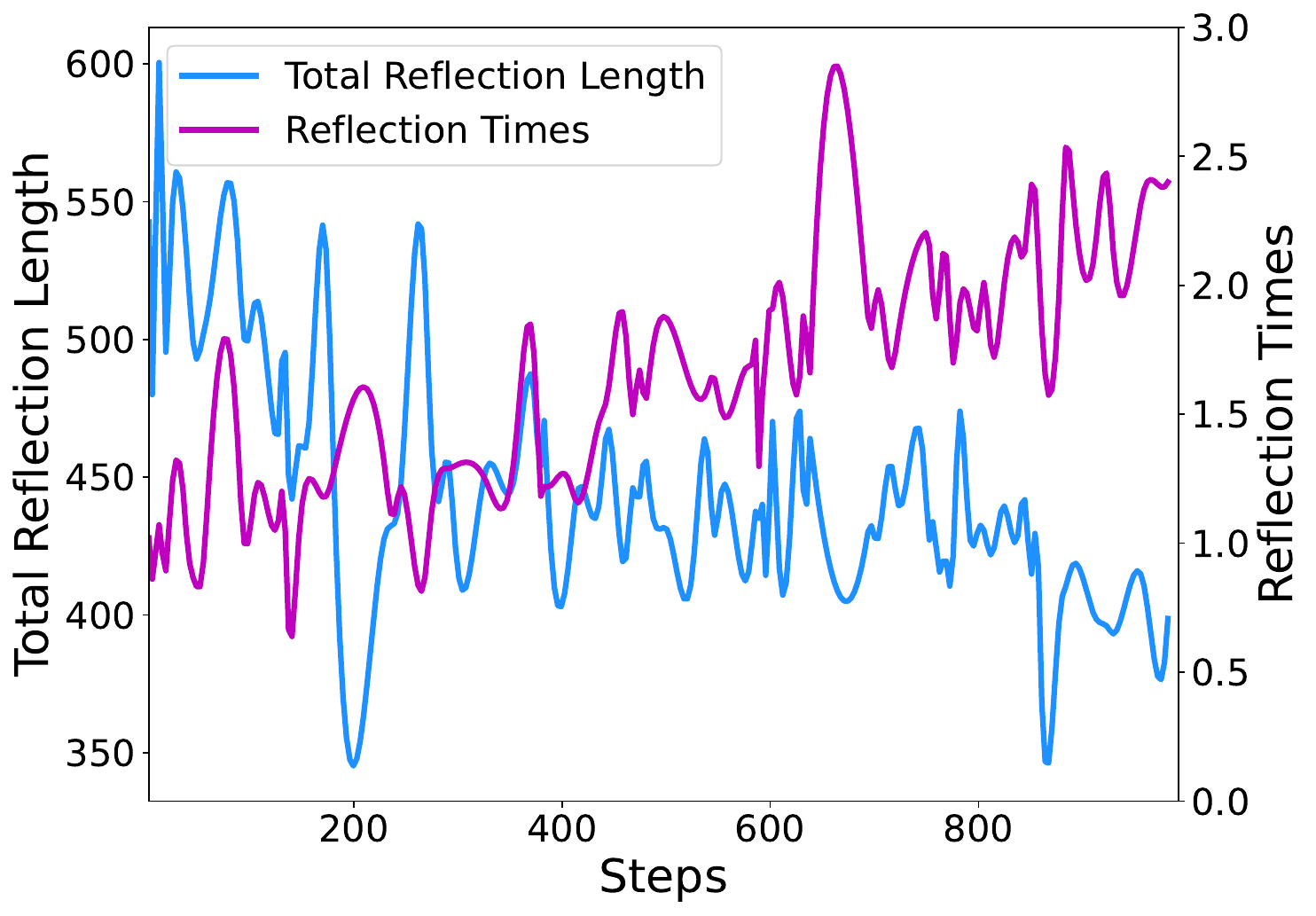}
        \caption{Total reflection length and reflection count of Qwen-Zero during the BRPO process. Qwen-Zero naturally learns to generate more frequent but more concise reflection processes.}
        \label{fig:length_with_times}
    \end{minipage}
    \hfill
    \begin{minipage}{0.48\textwidth}
        \centering
        \includegraphics[width=\textwidth]{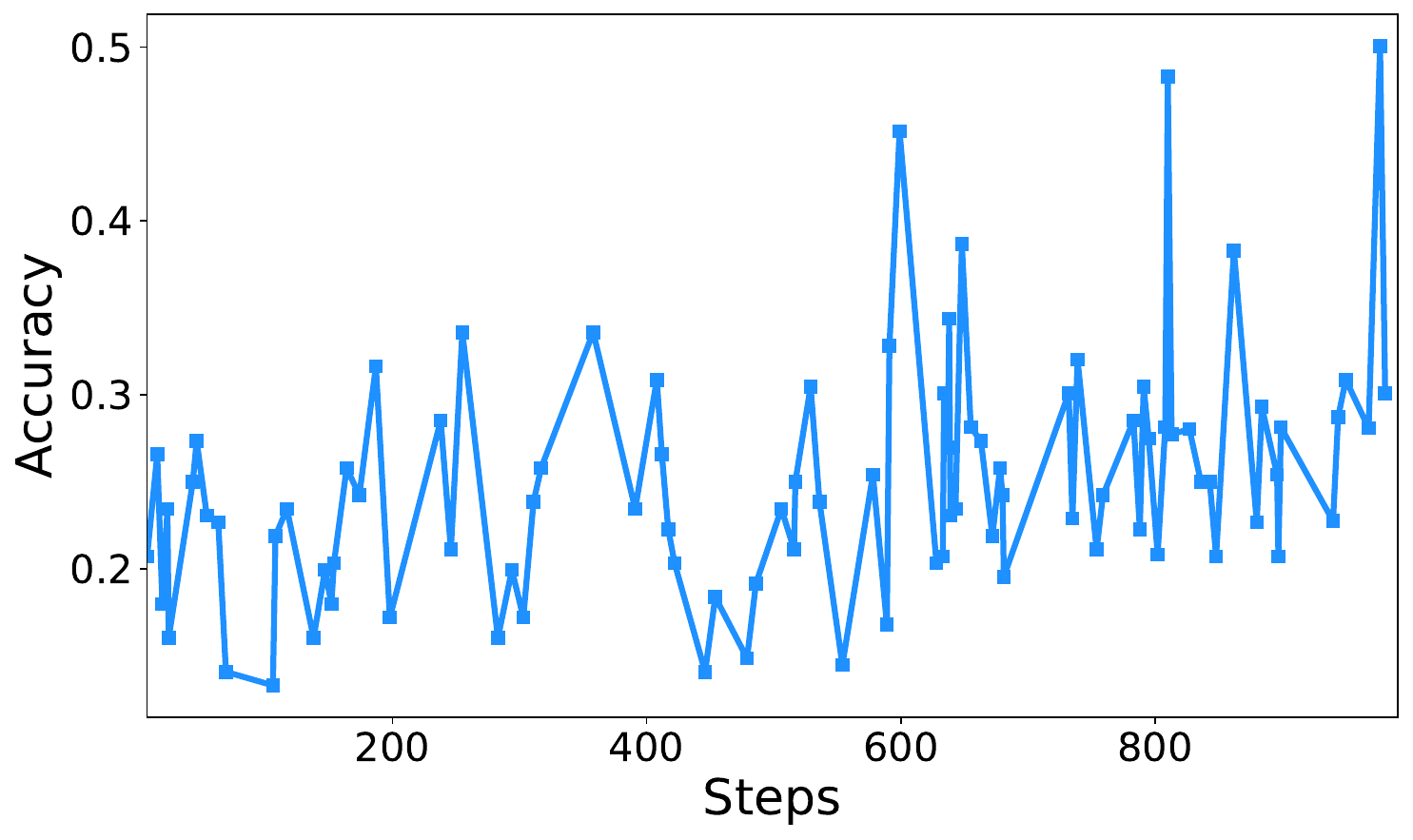}
        \caption{Changes in test accuracy during the BRPO process. Qwen-Zero gradually achieves higher accuracy as training progresses.}
        \label{fig:accuracy}
    \end{minipage}
\end{figure*}

As shown in Figure~\ref{fig:length_with_times}, during the BRPO training process, Qwen-Zero's average reflection length gradually decreases while the reflection count increases. The model naturally learns to generate more frequent but more concise reflection processes. Additionally, the test accuracy gradually increases during this process, as shown in Figure~\ref{fig:accuracy}.

\section{Cold Start Data Generation}
\begin{figure*}[h]
    \centering
    \includegraphics[width=\textwidth]{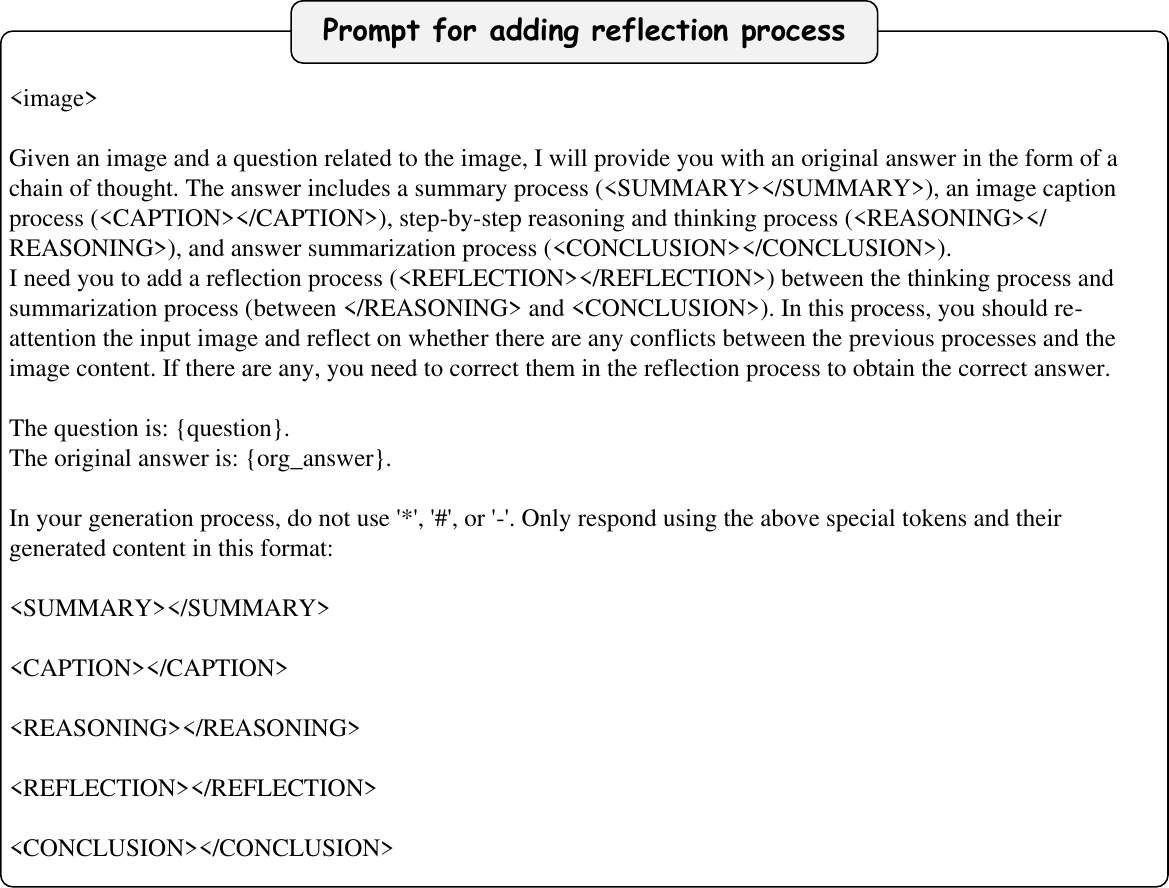}
    \caption{Prompt GPT-4o to generate cold start data with reflection process.}
    \label{fig:prompt_4o}
\end{figure*}

In section \ref{section:BRPO}, we propose using GPT-4o to construct cold-start data with reflection. Specifically, we extract 2k questions and answers from LLaVA-CoT-100k dataset~\cite{xu2411llava}, and prompt GPT-4o to insert a reflection process between reasoning and conclusion processes. The prompt we construct is shown in Figure~\ref {fig:prompt_4o}. The cold-start data underwent manual verification and correction, ensuring it does not contain ungrounded content or other factual errors. Each cold-start data item contains exactly one reflection.

\section{Detailed Proof}
\label{appendix:Detailed Proof}
\subsection{Proof of Theorem 3.1}
According to the chain rule, the generation entropy can be written as:
\begin{equation}
H(y \mid x,c)=\sum_{t=1}^{L_y} H\big(y_t \mid y_{<t}, x,c\big).
\end{equation}

Similarly, if ignoring the image prompt condition $c$:
\begin{equation}
H(y \mid x)=\sum_{t=1}^{L_y} H\big(y_t \mid y_{<t}, x\big).
\end{equation}

Therefore, the mutual information is:
\begin{equation}
I(y;c\mid x)=H(y\mid x)-H(y\mid x,c)
=\sum_{t=1}^{L_y} \Big[H\big(y_t\mid y_{<t},x\big)-H\big(y_t\mid y_{<t},x,c\big)\Big].
\end{equation}

Under the reasonable assumption that information from image prompt $c$ only enters the generation process in the form of $L_c$ tokens (i.e., no image content leaks in text prompt $x$), we can consider that throughout the generation process, each token receives contributions from $c$ approximately distributed in proportion $L_c/(L_x+L_c+L_y)$, that is:
\begin{equation}
I(y;c\mid x)\sim \frac{L_c}{L_{\text{total}}}\ H(y\mid x,c).
\end{equation}

Since $L_{\text{total}}=L_x+L_c+L_y$ increases as more tokens are generated, the ratio $r=\frac{L_c}{L_{\text{total}}}$ decreases, thus proving the theorem conclusion: during autoregressive generation, the model's attention to a fixed number of visual tokens decreases.

\subsection{Proof of Theorem 3.2}
Initially, the proportion of visual tokens is $r=\frac{L_c}{L_{\text{total}}}$. By copying or routing some visual tokens and inserting them into the generation sequence, the updated number of visual tokens and total length become $L_c+k$ and $L_{\text{total}}+k$ respectively, thus the new proportion is $r'=\frac{L_c+k}{L_{\text{total}}+k}$.

We need to prove $r' > r$, i.e.: $\frac{L_c + k}{L_{\text{total}} + k} > \frac{L_c}{L_{\text{total}}}$.

Cross-multiplying both sides:
\begin{equation}
(L_c + k) \cdot L_{\text{total}} > L_c \cdot (L_{\text{total}} + k).
\end{equation}

Further,
\begin{equation}
L_c \cdot L_{\text{total}} + k \cdot L_{\text{total}} > L_c \cdot L_{\text{total}} + L_c \cdot k.
\end{equation}

Canceling the common term $L_c \cdot L_{\text{total}}$ and dividing both sides by $k$ ($k > 0$):
\begin{equation}
L_{\text{total}} > L_c.
\end{equation}

Since $L_{\text{total}} = L_x + L_c + L_y$ (where $L_x$ is the number of text tokens and $L_y$ is the number of decoding text tokens), as long as $L_x +L_y > 0$, i.e., there exist any text tokens, then $L_{\text{total}} > L_c$, making $r' > r$ hold. Considering that in general cases, VLM input needs to include both text and image, therefore $L_x > 0$, meaning $L_{\text{total}} > L_c$ holds, thus proving $r' > r$.

Note that when $L_{\text{total}}$ is relatively large and $k>0$, $\frac{L_c+k}{L_{\text{total}}+k} > \frac{L_c}{L_{\text{total}}}$, making $I(y;c\mid x) \lesssim r'\cdot H(y\mid x,c)$ higher than the upper bound without additional visual tokens, therefore the model can achieve higher visual attention after increasing visual information.

\section{Training Details}
\label{appendix:Training Parameter Details}
We conduct full-parameter supervised fine-tuning of Qwen2.5-VL-Instruct on 8 NVIDIA A800-SXM4 80G GPUs. During training, the vision tower and multi-modal projector are frozen. Due to the addition of extra special tokens (e.g., <REFLECTION>), the embedding layer and head layer are partially unfrozen, optimizing only the weights of newly added special tokens. The training process employs DeepSpeed Zero-3 optimization strategy with a warmup ratio of 0.01. The learning rate is set to 1.0e-5, batch size is 1, gradient accumulation is 8, and training continues for 13 epochs. The image max pixels is set to 262144. We employ the AdamW optimizer with beta coefficients of 0.9 and 0.999, and an epsilon value of 1e-08.
\begin{figure*}[ht]
    \centering
    \begin{tabular}{cc}
        \includegraphics[width=0.48\textwidth]{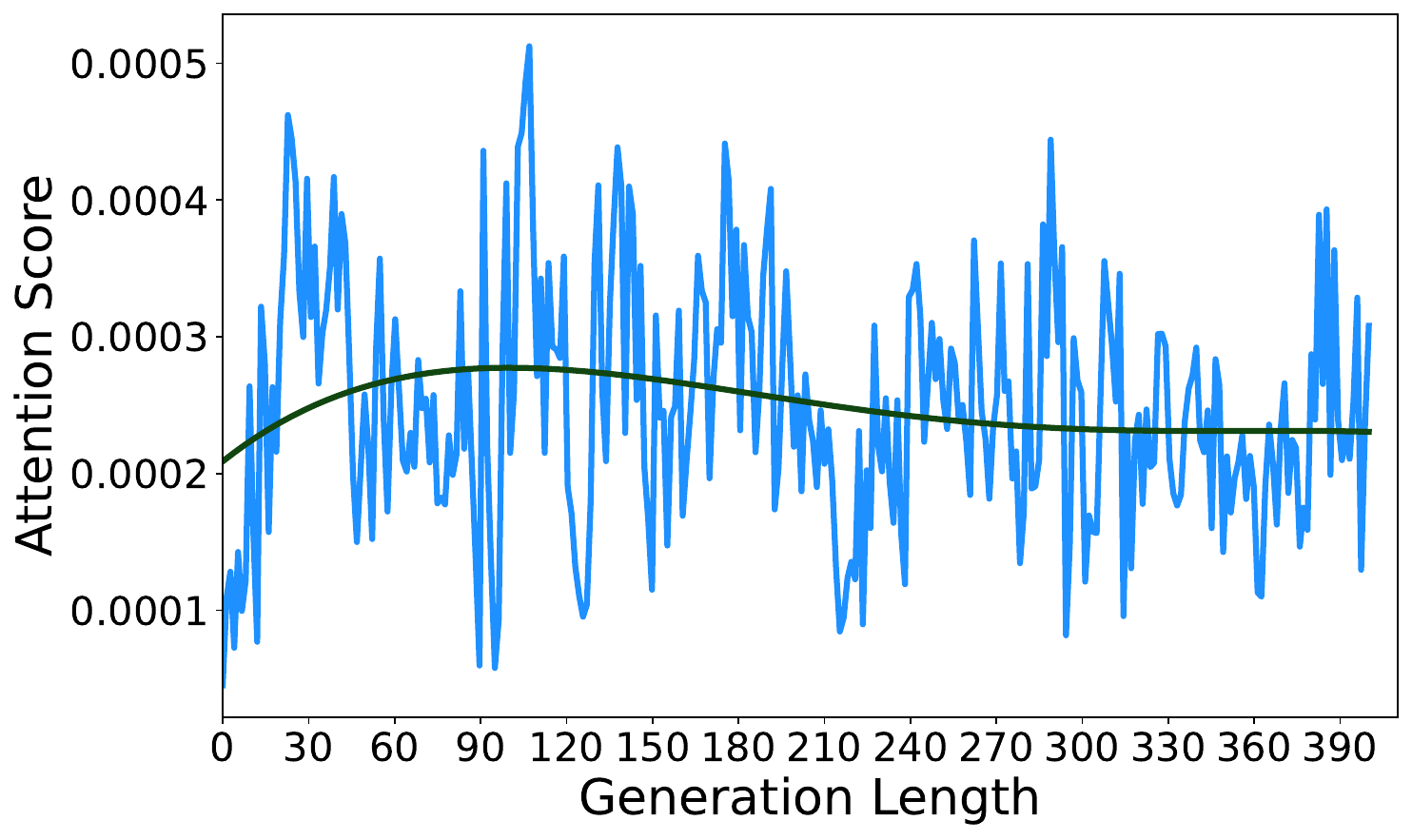} &
        \includegraphics[width=0.48\textwidth]{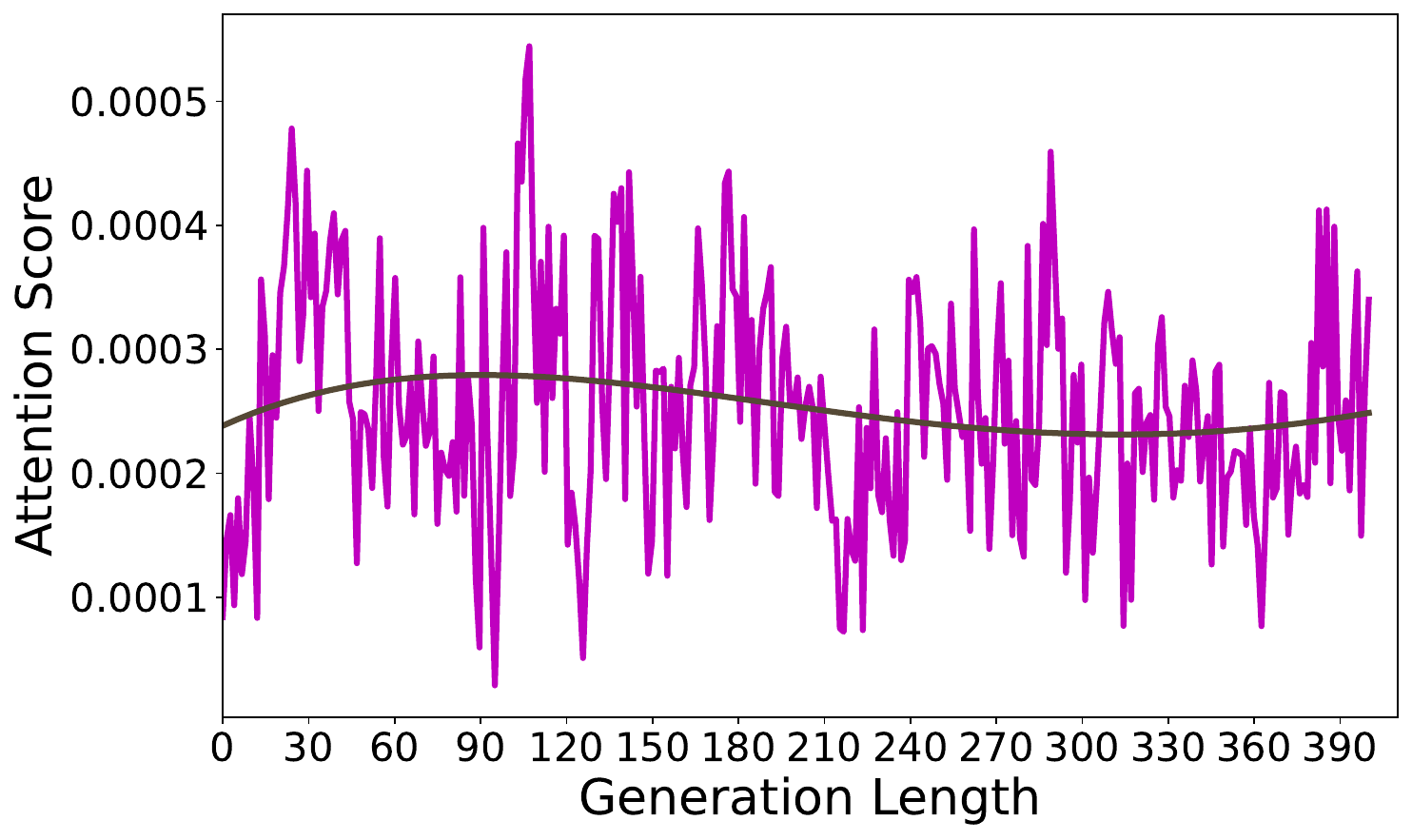}
    \end{tabular}
    \caption{Changes in average attention weights of visual tokens with increasing generation length for Qwen-LA-COPY (left) and Qwen-LA-ROUTE (right).}
    \label{fig:attn_vtc_vtr}
\end{figure*}

\section{Evaluating Vision-text Reflection}
We further compare the performance of text-only reflection, vision-only reflection, and our vision-text reflection (Qwen-LA-COPY and Qwen-LA-ROUTE), as shown in Table~\ref{tab:visiontext}. Text-only reflection means that when Qwen-LA performs reflection, it no longer copies or routes visual tokens, resulting in text-only reflection. Vision-only reflection means that when Qwen-LA-COPY performs reflection, it only copies visual tokens to the beginning of each reflection process and then forcibly ends the reflection (by inserting </REFLECTION>), resulting in vision-only reflection. Results show that compared to text-only and vision-only approaches, our vision-text reflection achieves higher accuracy, and hallucination metrics indicate that our vision-text reflection produces fewer hallucinations.

\begin{table*}[ht]
\centering
\caption{Performance comparison of Qwen-LA-COPY and Qwen-LA-ROUTE against text-only and vision-only reflection.}
\label{tab:visiontext}
\begin{tabular}{l|cccc}
\toprule
\textbf{Method} & \textbf{MMMU} $\uparrow$ & \textbf{MMStar} $\uparrow$ & \textbf{CHAIR}$_\text{i}$ $\downarrow$ & \textbf{MME} $\uparrow$ \\ \midrule
\textbf{Text-only} & 58.8 & 64.2 & 8.7 & 2308.5 \\
\textbf{Vision-only} & 56.1 & 63.3 & 8.4 & 2312.1 \\ \midrule
\textbf{Qwen-LA-COPY} & \textbf{60.3} & \textbf{65.9} & \textbf{3.7} & \textbf{2330.8} \\
\textbf{Qwen-LA-ROUT} & \underline{59.1} & \underline{64.6} & \underline{5.6} & \underline{2322.6} \\ \bottomrule
\end{tabular}
\end{table*}

\section{Data Validation}
\label{appendix:data validation}
In this paper, we introduce powerful models and human interventions to address various data quality issues, including cold-start data generated by GPT-4o and reasoning data generated by Qwen-Zero. Considering that most problems in our study are mathematical, we invite four PhD researchers to implement a rigorous manual verification and correction process for our dataset. Given the large volume of data, we adopt a two-stage approach: automatic model validation followed by human verification and correction for data flagged as incorrect by models.

During the automatic model validation phase, we use Claude-3.7\footnote{Claude 3.7 Sonnet, Anthropic, \url{www.anthropic.com/claude/sonnet}}, Gemini-2.5\footnote{Gemini 2.5 Pro, Google AI, \url{deepmind.google/technologies/gemini/}}, GPT-4o~\cite{jaech2024openai}, and Qwen2.5-VL-32B-Instruct~\cite{bai2025qwen2} to independently evaluate each answer. We instruct the models to check the accuracy of the answers, the coherence and correctness of the reasoning steps, and to output either "correct" or "incorrect." If two or more models judge an answer to be incorrect, we use GPT-4o to rewrite the reasoning process and answer, then apply human verification and correction.

All problems requiring verification are divided equally into four groups and assigned to the four PhD researchers for verification and correction. This process ensures the final dataset's high applicability while meeting rigorous mathematical standards.

\section{Broader Impacts}
\label{appendix:Ethical Considerations}
Our approach can advance research in the vision domain, especially in the field of Visual-Language Reasoning Models (VLRMs), promoting progress in model reasoning capabilities. Our Qwen-LA is able to suppress hallucinations in VLRMs and improve accuracy, creating positive impacts across various domains including visual question answering, robot control, and autonomous driving.
While our method does not directly introduce significant societal impacts, we acknowledge potential concerns regarding misuse of vision-language models. Malicious actors could potentially exploit our approach to generate illegal or harmful content, such as disinformation, deepfakes, or offensive material. Additionally, unauthorized use of our datasets or methodologies could raise privacy and security concerns. Though our work is primarily focused on fundamental research advancements rather than specific applications, we recognize the importance of responsible development and deployment. We encourage the research community to implement appropriate safeguards, such as content filtering mechanisms, usage monitoring, and ethical guidelines when building upon our work. We remain committed to open and transparent research while advocating for practices that minimize potential harms associated with vision-language technologies.

\section{Case Study}
\begin{figure*}[ht]
    \centering
    \includegraphics[width=\textwidth]{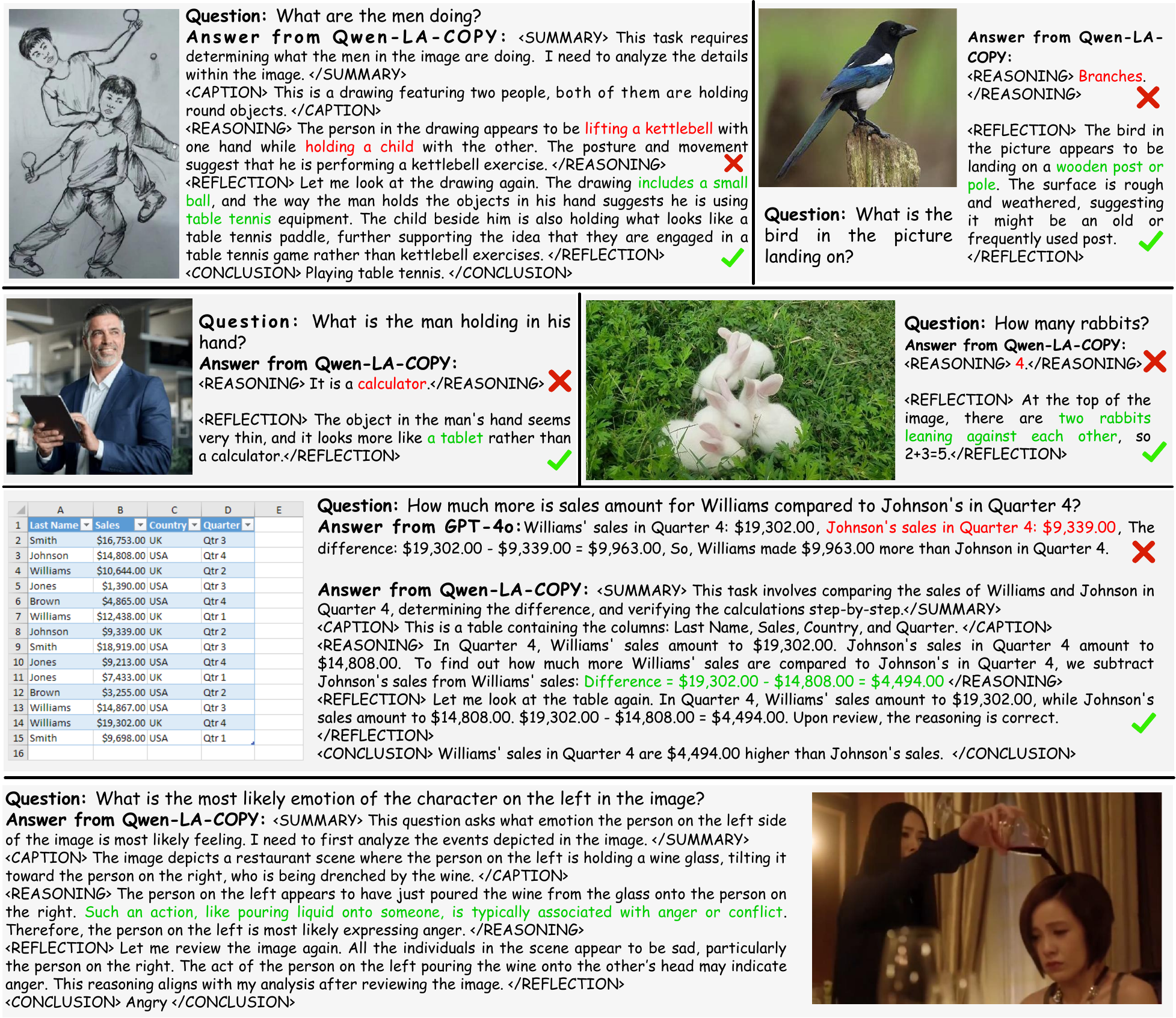}
    \caption{Example outputs of Qwen-LA-COPY. As shown in the first four examples, the reflection process can correct errors that occurred during the reasoning process. The fifth example shows that Qwen-LA-COPY completed the complex reasoning required for a table task, whereas GPT-4o failed. As shown in the sixth example, Qwen-LA-COPY can reach conclusions through reasoning. Even if the characters in the image do not display obvious facial expressions, Qwen-LA-COPY infers emotions by analyzing their actions.}
    \label{fig:example_all}
\end{figure*}

\end{document}